\documentclass[sigconf]{acmart}

\copyrightyear{2023} 
\acmYear{2023} 
\setcopyright{acmlicensed}\acmConference[KDD '23]{Proceedings of the 29th
ACM SIGKDD Conference on Knowledge Discovery and Data Mining}{August
6--10, 2023}{Long Beach, CA, USA}
\acmBooktitle{Proceedings of the 29th ACM SIGKDD Conference on Knowledge
Discovery and Data Mining (KDD '23), August 6--10, 2023, Long Beach, CA,
USA}
\acmPrice{15.00}
\acmDOI{10.1145/3580305.3599345}
\acmISBN{979-8-4007-0103-0/23/08}

\usepackage{graphicx}
\usepackage{subfigure}
\usepackage{caption}
\usepackage{makecell}
\usepackage{booktabs}
\usepackage{bm}
\usepackage{xcolor}
\usepackage{float}
\usepackage{multirow}
\usepackage{longtable}
\usepackage{framed}
\usepackage{enumitem}
\usepackage{amsthm,amsmath}
\usepackage{pythonhighlight}
\usepackage{appendix}
\usepackage[utf8]{inputenc}
\usepackage{bbding}
\usepackage{tabularx}
\usepackage{dsfont}

\DeclareMathAlphabet{\mathpzc}{OT1}{pzc}{m}{it}

\DeclareMathOperator*{\argmin}{arg\,min}
\usepackage{mathtools}
\usepackage{algorithm}
\usepackage{algpseudocode}
\usepackage{subfigure}
\usepackage{color,soul}

\usepackage{xspace}
\def\Method{FedCP\xspace}
\def\Policy{CPN\xspace}
\def\Policys{CPNs\xspace}

\makeatletter
\DeclareRobustCommand\onedot{\futurelet\@let@token\@onedot}
\def\@onedot{\ifx\@let@token.\else.\null\fi\xspace}
\def\eg{\emph{e.g}\onedot} 
\def\ie{\emph{i.e}\onedot} 
 
\def\etc{\emph{etc}\onedot}

\makeatother

\usepackage[utf8]{inputenc} 
\usepackage[T1]{fontenc}    
\usepackage{url}            
\usepackage{booktabs}       
\usepackage{amsfonts}       
\usepackage{nicefrac}       
\usepackage{microtype}      

\usepackage{hyperref}
\usepackage[capitalize]{cleveref}

\definecolor{blue_}{RGB}{76, 114, 176}
\definecolor{orange_}{RGB}{221, 132, 82}
\definecolor{upload}{RGB}{47, 85, 151}
\definecolor{download}{RGB}{241, 13, 208}
\definecolor{red_}{RGB}{255, 0, 0}
\definecolor{gray_}{RGB}{127, 127, 127}
\definecolor{green_}{RGB}{1, 128, 0}

\definecolor{blue__}{RGB}{65, 110, 152}
\definecolor{pink_}{RGB}{234, 99, 165}
\definecolor{purple_}{RGB}{154, 114, 242}

\usepackage{wrapfig}
\usepackage{balance}

\begin{document}

\title{\Method: Separating Feature Information for Personalized Federated Learning via Conditional Policy}


\author{Jianqing Zhang}
\affiliation{%
  \institution{Shanghai Jiao Tong University}
  \city{Shanghai}
  \country{China}}
\email{tsingz@sjtu.edu.cn}

\author{Yang Hua}
\affiliation{%
  \institution{Queen's University Belfast}
  \city{Belfast}
  \country{UK}}
\email{y.hua@qub.ac.uk}

\author{Hao Wang}
\affiliation{%
  \institution{Louisiana State University}
  \city{Baton Rouge}
  \country{USA}}
\email{haowang@lsu.edu}

\author{Tao Song}
\affiliation{%
  \institution{Shanghai Jiao Tong University}
  \city{Shanghai}
  \country{China}}
\email{songt333@sjtu.edu.cn}

\author{Zhengui Xue}
\affiliation{%
  \institution{Shanghai Jiao Tong University}
  \city{Shanghai}
  \country{China}}
\email{zhenguixue@sjtu.edu.cn}

\author{Ruhui Ma}
\affiliation{%
  \institution{Shanghai Jiao Tong University}
  \city{Shanghai}
  \country{China}}
\email{ruhuima@sjtu.edu.cn}

\author{Haibing Guan}
\affiliation{%
  \institution{Shanghai Jiao Tong University}
  \city{Shanghai}
  \country{China}}
\email{hbguan@sjtu.edu.cn}

\renewcommand{\shortauthors}{Jianqing Zhang et al.}

\begin{abstract}
Recently, personalized federated learning (pFL) has attracted increasing attention in privacy protection, collaborative learning, and tackling statistical heterogeneity among clients, \eg, hospitals, mobile smartphones, \etc. Most existing pFL methods focus on exploiting the global information and personalized information in the client-level model parameters while neglecting that data is the source of these two kinds of information. To address this, we propose the \textbf{Federated Conditional Policy (\Method)} method, which generates a conditional policy for each sample to separate the global information and personalized information in its features and then processes them by a global head and a personalized head, respectively. \Method is more fine-grained to consider personalization in a sample-specific manner than existing pFL methods. Extensive experiments in computer vision and natural language processing domains show that \Method outperforms eleven state-of-the-art methods by up to 6.69\%. Furthermore, \Method maintains its superiority when some clients accidentally drop out, which frequently happens in mobile settings. Our code is public at \url{https://github.com/TsingZ0/FedCP}.
\end{abstract}

\ccsdesc[500]{Computing methodologies~Multi-agent systems}
\ccsdesc[500]{Computing methodologies~Distributed algorithms}
\ccsdesc[300]{Computing methodologies~Supervised learning}

\keywords{Federated Learning; Statistical Heterogeneity; Personalization; Conditional Computing; Feature Separation}

\maketitle

\section{Introduction}
\label{sec:intro}
Nowadays, many web-based services, such as personalized recommendations~\cite{zhang2019deep, zhang2021tlsan, zhang2023lightfr}, benefit from artificial intelligence (AI) and the huge volume of data generated locally on various clients~\cite{kairouz2019advances}, \eg, hospitals, mobile smartphones, internet of things, \etc. At the same time, legislation endeavors on data privacy protection continue to increase, \eg, General Data Protection Regulation (GDPR) of Europe~\cite{regulation2016regulation} and California Consumer Privacy Act (CCPA)~\cite{de2018guide}. Due to privacy concerns and regulations, centralized AI faces significant challenges~\cite{nguyen2021federated, yang2020federated}. On the other hand, because of the data sparsity problem, it is hard to learn a reasonable model for a given task independently on each client~\cite{tan2022towards, kairouz2019advances, li2020federated}. 

Federated learning (FL) is proposed as a collaborative learning paradigm~\cite{mcmahan2017communication, kairouz2019advances, zhang2022fedala, ye2023feddisco} to utilize local data on the participating clients for the global model training without sharing the private data of clients. As one of the famous FL methods, FedAvg conducts four steps in each communication iteration: (1) The server sends the old global model parameters to the selected clients. (2) Each selected client initializes the local model with the received global parameters and trains the local model on local data. (3) The selected clients upload the updated local model parameters to the server. (4) The server generates new global model parameters by aggregating the received client model parameters. However, in practice, the data on the client is typically not independent and identically distributed (non-IID) as well as unbalanced~\cite{kairouz2019advances, li2020federated, zhang2022fedala, yang2019federated}. With this statistical heterogeneity challenge~\cite{li2020federated, tan2022towards}, the single global model in traditional FL methods, such as FedAvg, can hardly fit the local data well on each client and achieve good performance~\cite{huang2021personalized, t2020personalized}. 

\begin{figure}[h]
	\centering
	\includegraphics[width=\linewidth]{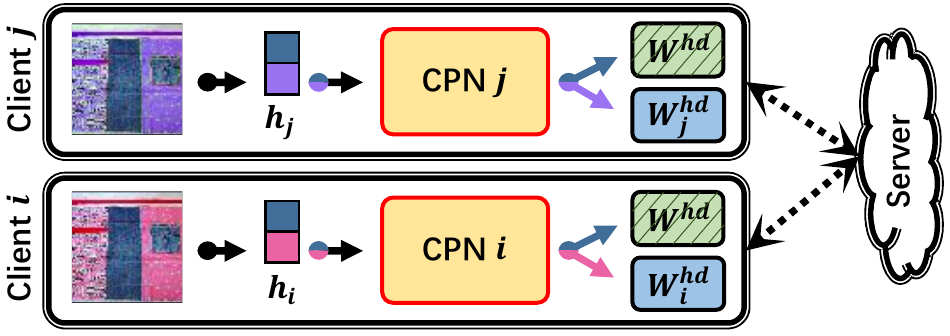}
	\caption{An example for \Method. ${\bm h}_{i/j}$: extracted feature vector, \Policy $i/j$: Conditional Policy Network, ${\bm W}^{hd}$: frozen global head, ${\bm W}^{hd}_{i/j}$: personalized head. Best viewed in color. }
	\label{fig:intro}
\end{figure}

To meet the personalized demand of each client and address the challenge of statistical heterogeneity in FL, personalized federated learning (pFL) comes along that focuses on learning personalized models rather than a single global model~\cite{t2020personalized, li2021fedphp}. Most existing pFL methods consider the global model as a container that stores the global information and enriches the personalized models with the parameters in the global model. 
However, they only focus on client-level model parameters, \ie, the global/personalized model to exploit the global/personalized information. Specifically, the meta-learning-based methods (such as Per-FedAvg~\cite{NEURIPS2020_24389bfe}) only fine-tune global model parameters to fit local data, and the regularization-based methods (such as pFedMe~\cite{t2020personalized}, FedAMP~\cite{huang2021personalized}, and Ditto~\cite{li2021ditto}) only regularize model parameters during local training. 
Although personalized-head-based methods (such as FedPer\cite{arivazhagan2019federated}, FedRep~\cite{collins2021exploiting}, and FedRoD~\cite{chen2021bridging}) explicitly split a backbone into a global part (feature extractor) and a personalized part (head), they still focus on exploiting global and personalized information in model parameters rather than the source of information: \textbf{\textit{data}}. 
As the model is trained on data, \textbf{the global/personalized information in model parameters is derived from client data}. In other words, the heterogeneous data on clients contains both global and personalized information. As shown in \Cref{fig:intro}, widely-used colors, \eg, \textcolor{blue__}{blue}, and rarely-used colors, \eg, \textcolor{purple_}{purple} and \textcolor{pink_}{pink}, contain global information and personalized information in images, respectively. 

To exploit the global and personalized information in the data separately, we propose a \textbf{Federated Conditional Policy (\Method)} method based on conditional computing techniques~\cite{guo2019spottune, oreshkin2018tadam}. Since the dimension of raw input data is much larger than the feature vector extracted by the feature extractor, we focus on the feature vector for efficiency. As the proportion of the global and personalized information in the features differ among samples and clients, we propose an auxiliary \textbf{Conditional Policy Network (\Policy)} to generate the sample-specific policy for feature information separation. Then, we process the global feature information and personalized feature information by a global head and a personalized head in different routes, respectively, as shown in \Cref{fig:intro}. We store the personalized information in the personalized head and reserve the global information by freezing the global head without locally training it. 
Through end-to-end learning, \Policy automatically learns to generate the sample-specific policy. We visualize six cases in \Cref{sec:act} to show the effectiveness of the feature information separation ability. 

To evaluate \Method, we conduct extensive experiments on various datasets in two widely-used scenarios~\cite{mcmahan2017communication, li2021model}, \ie, the pathological settings and the practical settings. \Method outperforms eleven state-of-the-art (SOTA) methods in both scenarios, and we analyze the reasons in \Cref{sec:main_exp}. 
In summary, our key contributions are: 

\begin{itemize}
    \item To the best of our knowledge, we are the first to consider personalization on the sample-specific feature information in FL. It is more fine-grained than using the client-level model parameters in most existing FL methods. 
    \item We propose a novel \Method that generates a sample-specific policy to separate the global information and personalized information in features on each client. It processes these two kinds of feature information through a frozen global head and a personalized head on each client, respectively. 
    \item We conduct extensive experiments in computer vision (CV) and natural language processing (NLP) domains to show the effectiveness of \Method. Besides, \Method keeps its superior performance even when some clients accidentally drop out.
\end{itemize}

\section{Related Work}
\label{sec:related}

\subsection{Personalized Federated Learning}
To collaboratively learn models among clients on their local private data while protecting privacy, traditional FL methods, such as FedAvg~\cite{mcmahan2017communication} and FedProx~\cite{MLSYS2020_38af8613}, come along. Based on FedAvg, FedProx improves the stability of the FL process through a regularization term. However, in practice, statistical heterogeneity widely exists in the FL setting, so it is hard to learn a single global model that fits well with the local data in each client~\cite{kairouz2019advances, huang2021personalized, t2020personalized}. 

Recently, pFL has attracted increasing attention for its ability to tackle statistical heterogeneity in FL~\cite{kairouz2019advances, hahn2022connecting}. Among \textbf{meta-learning-based methods}, Per-FedAvg~\cite{NEURIPS2020_24389bfe} learns an initial shared model as the global model that satisfies the learning trend for each client. 
Among \textbf{regularization-based methods}, pFedMe~\cite{t2020personalized} learns an additional personalized model locally for each client with Moreau envelopes. In addition to learning only one global model for all clients, FedAMP~\cite{huang2021personalized} generates one server model for one client through the attention-inducing function to find similar clients. In Ditto~\cite{li2021ditto}, each client learns its personalized model locally with a proximal term to fetch global information from global model parameters. Among \textbf{personalized-head-based methods}, FedPer\cite{arivazhagan2019federated} and FedRep~\cite{collins2021exploiting} learn a global feature extractor and a client-specific head. The former locally trains the head with the feature extractor, while the latter locally fine-tunes the head until convergence before training the feature extractor in each iteration. To bridge traditional FL and pFL, FedRoD~\cite{chen2021bridging} explicitly learns two prediction tasks with a global feature extractor and two heads. It uses the balanced softmax (BSM) loss~\cite{ren2020balanced} for the global prediction task and processes the personalized task by the personalized head. Among \textbf{other pFL methods}, FedFomo~\cite{zhang2020personalized} calculates the client-specific weights for aggregation on each client using the personalized models from other clients. FedPHP~\cite{li2021fedphp} locally aggregates the global model and the old personalized model using a moving average to keep the historical personalized information. It also transfers the information in the global feature extractor through the widely-used maximum mean discrepancy (MMD) loss~\cite{gretton2006kernel, qin2019pointdan}. 
These above pFL methods only focus on exploiting global and personalized information of model parameters but do not dig deep into data. 

\begin{figure*}[h]
	\centering
	\subfigure[Forward data flow corresponding to the local learning on client $i$.]{\includegraphics[width=0.57\linewidth]{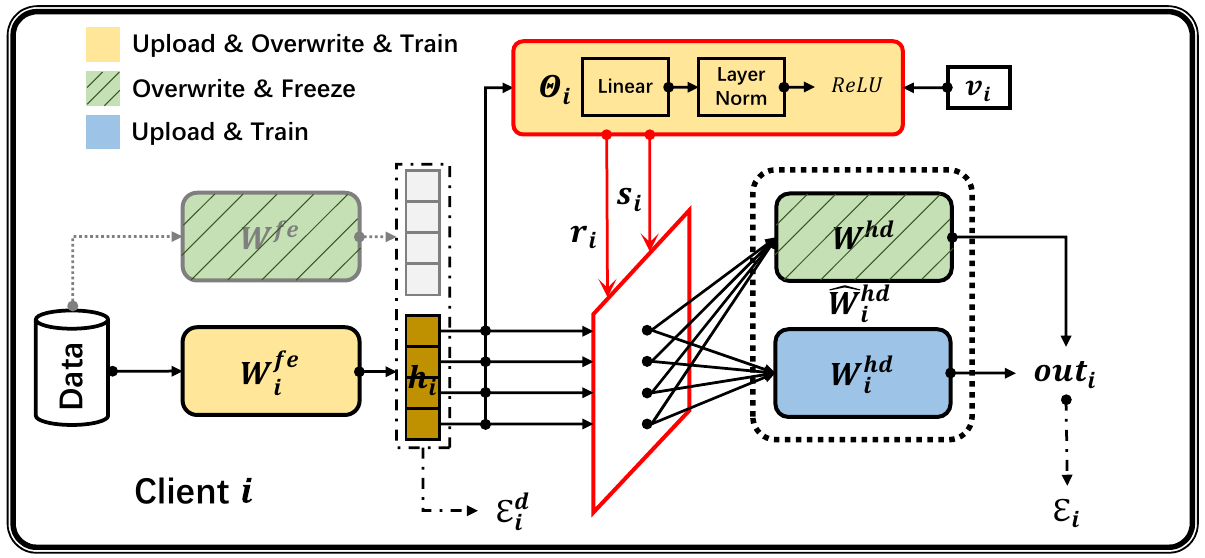}\label{fig:cp_a}}
    \hfill
	\subfigure[\textcolor{upload}{Upload} and \textcolor{download}{download} streams in \Method.]{\includegraphics[width=0.42\linewidth]{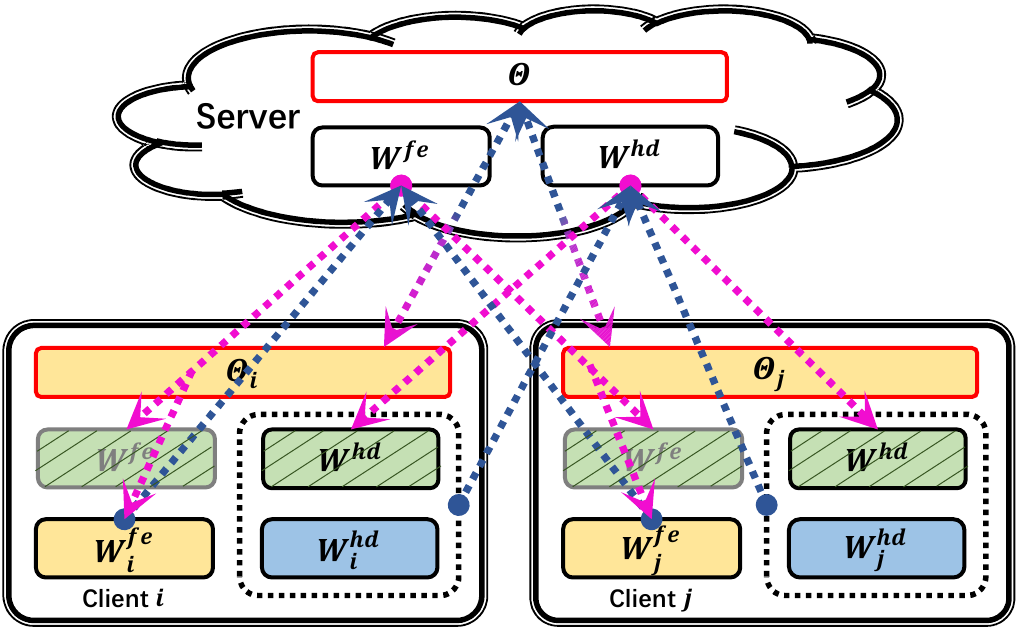}\label{fig:cp_b}}
	\caption{(a) The conditional policy separates information from ${\bm h}_i$ into ${\bm r}_i \odot {\bm h}_i$ and ${\bm s}_i \odot {\bm h}_i$ in the \textcolor{red}{red rhomboid}. Except for the feature vectors and vector ${\bm v}_i$, a standard rectangle and a rounded rectangle represent a layer and a module, respectively. The rounded rectangle with the dashed border is $\widehat{{\bm W}}^{hd}_i$ in \cref{eq:3}. ${\bm W}^{fe}$ (\textcolor{gray_}{gray border}) is not a part of the personalized model, where data only flows forward during training. Data flows in all the lines \emph{during training}, but it only flows in the solid lines \emph{during inference}. (b) For clarity, we separately show the \textcolor{upload}{upload} and \textcolor{download}{download} streams for the feature extractors, the heads, and the \Policys. Still, we upload or download them as a union between the server and each client in practice. Best viewed in color. }
	\label{fig:cp}
\end{figure*}

\subsection{Conditional Computing}

Conditional computing is a technique that introduces dynamic characteristics into models according to task-dependent conditional inputs~\cite{liu2018dynamic, guo2019spottune, oreshkin2018tadam}. Formally, given a conditional input $C$ (\eg, image/text, model parameter vector, or other auxiliary information) and an auxiliary module $AM(\cdot; \theta)$, a signal $S$ can be generated by $S = AM(C; \theta)$ and used to interfere with models, such as dynamic routing and feature adaptation. 

To activate specific parts in a model and process the data in different routes for each input sample, many approaches generate sample-specific policies for route selection. Conditioned on the input image, ConvNet-AIG~\cite{veit2018convolutional} can decide which layers are needed during inference using Gumbel Softmax~\cite{jang2016categorical}. 
With a policy network, SpotTune~\cite{guo2019spottune} makes decisions for each image to select which blocks in a pre-trained residual network to fine-tune. 

Instead of focusing on dynamic model topology, some methods propose adapting the learned features. In the few-shot learning field, TADAM~\cite{oreshkin2018tadam} adapts the features through an affine transformation conditioned by the extracted task representation. In the video object detection field, TMA~\cite{hua2021temporal} proposes a learnable affine transformation conditioned by video frames for feature adaptation. 

The above methods use conditional computing techniques but are designed for centralized AI scenarios and specific tasks. Combining the ideas of dynamic routing and feature adaptation, we devise the \Policy module in our \Method to separate global feature information and personalized feature information then process them in different routes for pFL scenarios and various tasks.

\section{Method}
\label{sec:method}

\subsection{Overview}
In statistically heterogeneous pFL settings, non-IID and unbalanced data exist on $N$ clients, who train their personalized models ${\bm W}_1, \ldots, {\bm W}_N$ in a collaborative manner. $N$ clients own private datasets $\mathcal{D}_1, \ldots, \mathcal{D}_N$, respectively, which are sampled from $N$ distinct distributions without overlapping. 

Similar to FedPer\cite{arivazhagan2019federated},  FedRep~\cite{collins2021exploiting}, and  FedRoD~\cite{chen2021bridging}, we split the backbone into a feature extractor $f:\mathbb{R}^D \rightarrow \mathbb{R}^K$, that maps input samples to feature space and a head $g:\mathbb{R}^K \rightarrow \mathbb{R}^C$, which maps from low-dimensional feature space to a label space. Following FedRep, we consider the last fully connected (FC) layer in each given backbone as the head. $D$, $K$, and $C$ are the dimension of the input space, feature space, and label space, respectively. $K$ is determined by the given backbone and typically $D \gg K$. 

Different from FedPer, FedRep and FedRoD, on client $i$, we have a global feature extractor (parameterized by ${\bm W}^{fe}$), a global head (parameterized by ${\bm W}^{hd}$), a personalized feature extractor (parameterized by ${\bm W}^{fe}_i$), a personalized head (parameterized by ${\bm W}^{hd}_i$), and a \Policy (parameterized by ${\bm \Theta}_i$). 
Specifically, \textbf{for the feature extractors}, we initialize ${\bm W}^{fe}_i$ by overwriting it with corresponding global parameters ${\bm W}^{fe}$ in each iteration, and then locally learn the personalized feature extractor. The feature generated by the changing personalized feature extractor may not fit the frozen global head during local learning. Thus, we freeze the global feature extractor after receiving and align the features outputted by the personalized feature extractor to the ones generated by the global feature extractor through the MMD loss, as shown in \Cref{fig:cp_a}. \textbf{For the global head}, we freeze it after it has been initialized by ${\bm W}^{hd}$ to preserve global information. 
In short, at the start of each iteration, we overwrite ${\bm W}^{fe}_i$ by new ${\bm W}^{fe}$ then freeze ${\bm W}^{fe}$ and ${\bm W}^{hd}$. 
As shown by the non-transparent module in \Cref{fig:cp_a}, the personalized model used for inference (parameterized by ${\bm W}_i$) consists of the personalized feature extractor, the global head, the personalized head, and the \Policy, \ie, ${\bm W}_i := \{{\bm W}^{fe}_i, {\bm W}^{hd}, {\bm W}^{hd}_i, {\bm \Theta}_i\}$. The frozen global feature extractor is only used for local learning and is not part of the personalized model. We omit iteration notation, sample index notation, and biases for simplicity. Given the local loss $\mathcal{F}_i$ (described later), our objective is
\begin{equation}
    \{{\bm W}_1, \ldots, {\bm W}_N\} = \argmin \ \mathcal{G}(\mathcal{F}_1, \ldots, \mathcal{F}_N).
\end{equation}
Typically, $\mathcal{G}(\mathcal{F}_1, \ldots, \mathcal{F}_N) = \sum^{N}_{i=1} n_i \mathcal{F}_i$, $n_i = |\mathcal{D}_i| / \sum^{N}_{j=1} |\mathcal{D}_j|$, and $|\mathcal{D}_i|$ is the sample amount on client $i$.

\subsection{Federated Conditional Policy (\Method)}
\label{sec:fcp}

We focus on feature information separation for the feature vector
\begin{equation}
    {\bm h}_i = f({\bm x}_i; {\bm W}^{fe}_i), \forall ({\bm x}_i, y_i) \in \mathcal{D}_i. \label{eq:feat}
\end{equation}
Due to statistical heterogeneity, ${\bm h}_i \in \mathbb{R}^K$ contains global and personalized feature information. To separately exploit these two kinds of information, we propose \textbf{\Method} that learns sample-specific separation in an end-to-end manner, as shown in \Cref{fig:cp}. 

\subsubsection{Separating feature information} Guided by the global information in the frozen global head and the personalized information in the personalized head, the \Policy (the core of \Method) can learn to generate the sample-specific policy and separate the global and personalized information in ${\bm h}_i$ automatically. 

Specifically, we devise \Policy as the concatenation of an FC layer and a layer-normalization layer~\cite{ba2016layer} followed by the ReLU activation function~\cite{li2017convergence}, as shown in \Cref{fig:cp_a}. On client $i$, we generate the sample-specific policy by
\begin{equation}
    \{{\bm r}_i, {\bm s}_i\} := {\rm \Policy}(\mathcal{C}_i; {\bm \Theta}_i), \label{eq:policy}
\end{equation}
where ${\bm r}_i\in \mathbb{R}^K, {\bm s}_i\in \mathbb{R}^K, r^k_{i} + s^k_{i} = 1, \forall k \in [K]$, and $\mathcal{C}_i \in \mathbb{R}^K$ is the sample-specific input for \Policy. We describe the details of the input $\mathcal{C}_i$ and the output $\{{\bm r}_i, {\bm s}_i\}$ as follows. 

$\mathcal{C}_i$ is generated to achieve the sample-specific characteristic and introduce personalized (client-specific) information. We can directly obtain the sample-specific vector ${\bm h}_i$, so we only introduce how to obtain the client-specific information here. 
Based on FedRep and FedRoD, the parameters in the personalized head, \ie, ${\bm W}^{hd}_i$, naturally contain client-specific information. However, ${\bm W}^{hd}_i$ is a matrix, not a vector. Thus, we generate ${\bm v}_i$ by reducing the dimension of ${\bm W}^{hd}_i$. 
Recall that a head is an FC layer in \Method, \ie, ${\bm W}^{hd}_i \in \mathbb{R}^{C\times K}$, so the $k$th column of ${\bm W}^{hd}_i$ corresponds to $k$th feature in ${\bm h}_i$. We obtain ${\bm v}_i := \sum^{C}_{c=1} {\bm w}^T_c,$ where ${\bm w}_c$ is the $c$th row in ${\bm W}^{hd}_i$ and ${\bm v}_i \in \mathbb{R}^{K}$. In this way, we obtain a client-specific vector with the same shape and feature-wise semantics as ${\bm h}_i$. Then we combine sample-specific ${\bm h}_i$ and the client-specific ${\bm v}_i$ via
$\mathcal{C}_i:=({\bm v}_i / ||{\bm v}_i||_2) \odot {\bm h}_i$, where $||{\bm v}_i||_2$ is the $\ell_2$-norm~\cite{perronnin2010improving} of ${\bm v}_i$ and $\odot$ is the Hadamard product. We obtain ${\bm v}_i$ before local learning in each iteration and regard it as a constant during training. During inference, we reuse the latest ${\bm v}_i$.

We separate information by multiplying the policy $\{{\bm r}_i, {\bm s}_i\}$ and ${\bm h}_i$ to obtain the global feature information ${\bm r}_i \odot {\bm h}_i$ and personalized feature information ${\bm s}_i \odot {\bm h}_i$. There are connections among features~\cite{yu2003feature}, so we output $\{{\bm r}_i, {\bm s}_i\}$ with real numbers instead of Boolean values, \ie, $r^k_{i} \in (0, 1)$ and $s^k_{i} \in (0, 1)$. Inspired by the Gumbel-Max trick for policy generating~\cite{guo2019spottune}, we generate the policy with the help of the intermediates and a softmax~\cite{hinton2015distilling} operation through the following two steps. Firstly, \Policy generates the intermediates ${\bm a}_i \in \mathbb{R}^{K\times 2}$, where $a^k_{i} = \{a^{k}_{i, 1}, a^{k}_{i, 2}\}, k \in [K]$, $a^{k}_{i, 1}$ and $a^{k}_{i, 2}$ are scalars without constraint. 
Secondly, we obtain $r^k_{i}$ and $s^k_{i}$ by
\begin{equation}
    r^k_{i} = \frac{\exp{(a^{k}_{i, 1})}}{\sum_{j\in \{1, 2\}}\exp{(a^{k}_{i, j})}}, \quad s^k_{i} = \frac{\exp{(a^{k}_{i, 2})}}{\sum_{j\in \{1, 2\}}\exp{(a^{k}_{i, j})}}.
\end{equation}
Note that, $r^k_{i} \in (0, 1), s^k_{i} \in (0, 1), r^k_{i} + s^k_{i} = 1, \forall k \in [K]$ still holds. 

\begin{algorithm}[t]
	\caption{The Learning Process in \Method}
	\begin{algorithmic}[1]
		\Require 
		$N$ clients with their local data, 
		${\bm W}^{fe, 0}$: initial parameters of the global feature extractor, 
		${\bm W}^{hd, 0}$: initial parameters of the global head, 
		${\bm \Theta}^0$: initial parameters of the global \Policy, 
		$\eta$: local learning rate, 
		$\lambda$: hyper-parameter for MMD loss, 
		$\rho \in (0, 1]$: client joining ratio in one iteration, 
		$T$: total training iterations. 
		\Ensure 
		Reasonable personalized models $\{{\bm W}_1, \ldots, {\bm W}_N\}$.
		\State Server sends ${\bm W}^{fe, 0}$ and ${\bm W}^{hd, 0}$ to initialize ${\bm W}^{fe}$, ${\bm W}^{hd}$, ${\bm W}^{fe}_i$, \Statex \qquad and ${\bm W}^{hd}_i$ on client $i, \forall i\in [N]$. 
		\State Server sends ${\bm \Theta}^0$ to initialize the \Policy on client $i, \forall i\in [N]$. 
		\For{iteration $t=0, \ldots, T$}
		    \State Server randomly samples a subset $\mathcal{I}^t$ of clients based on $\rho$.
		    \State Server sends ${\bm W}^{fe, t}$, ${\bm W}^{hd, t}$, and ${\bm \Theta}^t$ to the selected clients. 
		    \For{Client $i \in \mathcal{I}^t$ in parallel}
		        \Statex \Comment{\textbf{local initialization}}
		        \State Client $i$ overwrites ${\bm W}^{fe}$ and ${\bm W}^{fe}_i$ with the parameters \Statex \qquad \qquad \quad ${\bm W}^{fe, t}$ and freezes ${\bm W}^{fe}$.
		        \State Client $i$ overwrites ${\bm W}^{hd}$ with the parameters ${\bm W}^{hd, t}$ \Statex \qquad \qquad \quad and freezes ${\bm W}^{hd}$. 
		        \State Client $i$ overwrites ${\bm \Theta}_i$ with the parameters ${\bm \Theta}^t$.
		        \State Client $i$ generates the client-specific vector ${\bm v}_i$.
		        \Statex \Comment{\textbf{local learning}}
		        \State Client $i$ updates ${\bm W}^{fe}_i$, ${\bm W}^{hd}_i$ and ${\bm \Theta}_i$ simultaneously:
		            \State \qquad ${\bm W}^{fe}_i \leftarrow {\bm W}^{fe}_i - \eta \nabla_{{\bm W}^{fe}_i} \mathcal{F}_i$; 
		            \State \qquad ${\bm W}^{hd}_i \leftarrow {\bm W}^{hd}_i - \eta \nabla_{{\bm W}^{hd}_i} \mathcal{F}_i$; 
		            \State \qquad ${\bm \Theta}_i \leftarrow {\bm \Theta}_i - \eta \nabla_{{\bm \Theta}_i} \mathcal{F}_i$.
		        \State Client $i$ obtains $\widehat{{\bm W}}^{hd}_i$ through \cref{eq:3}. 
		        \State Client $i$ uploads $\{{\bm W}^{fe}_i, \widehat{{\bm W}}^{hd}_i, {\bm \Theta}_i\}$ to the server. 
		    \EndFor
		    \Statex \Comment{\textbf{Server aggregation}}
		    \State Server calculates $n^t = \sum_{i \in
		    \mathcal{I}^t} n_i$ and obtains 
		        \State \qquad ${\bm W}^{fe, t+1} = \frac{1}{n^t} \sum_{i \in \mathcal{I}^t} n_i {\bm W}^{fe}_i$; 
		        \State \qquad ${\bm W}^{hd, t+1} = \frac{1}{n^t} \sum_{i \in \mathcal{I}^t} n_i \widehat{{\bm W}}^{hd}_i$; 
		        \State \qquad ${\bm \Theta}^{t+1} = \frac{1}{n^t} \sum_{i \in \mathcal{I}^t} n_i {\bm \Theta}_i$.
		\EndFor
		\\
		\Return $\{{\bm W}_1, \ldots, {\bm W}_N\}$
	\end{algorithmic}
	\label{algo}
\end{algorithm}

\subsubsection{Processing feature information}

Then, we feed ${\bm r}_i \odot{\bm h}_i$ and ${\bm s}_i \odot {\bm h}_i$ to the global head and the personalized head, respectively. The outputs of global head and the personalized head are ${\bm {out}^r_i} = g({\bm r}_i \odot{\bm h}_i; {\bm W}^{hd})$ and ${\bm {out}^s_i} = g({\bm s}_i \odot{\bm h}_i; {\bm W}^{hd}_i)$, respectively. We define the final output ${\bm {out}_i} := {\bm {out}^r_i} + {\bm {out}^s_i}$. Then the local loss is
\begin{equation}
    \mathcal{E}_i = \mathbb{E}_{({\bm x}_i, y_i) \sim \mathcal{D}_i} \mathcal{L}({\bm {out}_i}, y_i), \label{eq:e_loss}
\end{equation}
where $\mathcal{L}$ is the cross-entropy loss function~\cite{murphy2012machine}. 

From the view of each sample, the extracted features are processed by both the global head and the personalized head. For simplicity, we aggregate these two heads through averaging to form the upload head $\widehat{{\bm W}}^{hd}_i$:
\begin{equation}
    \widehat{{\bm W}}^{hd}_i = \frac{{\bm W}^{hd} + {\bm W}^{hd}_i}{2}. \label{eq:3}
\end{equation}
In each iteration, we upload $\{{\bm W}^{fe}_i, \widehat{{\bm W}}^{hd}_i, {\bm \Theta}_i\}$ to the server. 

\subsubsection{Aligning features}

To fit the features outputted by the personalized feature extractor with the frozen global head, we align the features outputted by the personalized feature extractor and the global feature extractor through the MMD loss $\mathcal{E}^{d}_i$,
\begin{equation}
    \mathcal{E}^{d}_i = ||\mathbb{E}_{({\bm x}_i, y_i) \sim \mathcal{D}_i} \phi({\bm h}_i) - \mathbb{E}_{({\bm x}_i, y_i) \sim \mathcal{D}_i} \phi(f({\bm x}_i; {\bm W}^{fe}))||_{\mathcal{H}}^2 ,
\end{equation}
where $\mathcal{H}$ is a reproducing kernel Hilbert space (RKHS) and $\phi$ is induced by a specific kernel function (\eg, the radial basis function (RBF)), \ie, $\kappa({\bm h}_i, {\bm h}_j) = \langle \phi({\bm h}_i), \phi({\bm h}_j) \rangle$~\cite{li2021fedphp}. Finally, we have the local loss $\mathcal{F}_i = \mathcal{E}_i + \lambda \mathcal{E}^{d}_i$, where $\lambda$ is a hyper-parameter. Specifically, 
\begin{equation}
\begin{aligned}
    \mathcal{F}_i &=\mathbb{E}_{({\bm x}_i, y_i) \sim \mathcal{D}_i} \mathcal{L}[g({\bm r}_i \odot{\bm h}_i; {\bm W}^{hd}) + g({\bm s}_i \odot{\bm h}_i; {\bm W}^{hd}_i), y_i] \\
    & + \lambda ||\mathbb{E}_{({\bm x}_i, y_i) \sim \mathcal{D}_i} \phi({\bm h}_i) - \mathbb{E}_{({\bm x}_i, y_i) \sim \mathcal{D}_i} \phi(f({\bm x}_i; {\bm W}^{fe}))||_{\mathcal{H}}^2, \label{eq:final}
\end{aligned}
\end{equation}
where ${\bm h}_i$ is the feature vector extracted by \cref{eq:feat}, and ${\bm r}_i$ and ${\bm s}_i$ are obtained through ~\cref{eq:policy}. We show the entire \emph{\textbf{learning}} process in \Cref{algo} and the model for \emph{\textbf{inference}} in \Cref{fig:cp_a}. 

\subsection{Privacy Analysis}

According to \Cref{fig:cp_b} and \Cref{algo}, our proposed \Method shares the parameters of one feature extractor, one head, and one \Policy. As for the head part, we upload $\widehat{{\bm W}}^{hd}_i$ on each client after aggregating ${\bm W}^{hd}$ and ${\bm W}^{hd}_i$ by \cref{eq:3}. This process can be viewed as adding noise (global parameters ${\bm W}^{hd}$) to ${\bm W}^{hd}_i$, thus protecting privacy during the uploading and downloading. Besides, the sample-specific characteristic further improves the privacy-preserving ability of \Method. On the one hand, since $\mathcal{C}_i$ is dynamically generated without sharing with the server, it is hard to recover the sample-specific policy with the \Policy or through model inversion attacks~\cite{al2016reconstruction}. On the other hand, without the sample-specific policy, the connection between the feature extractor and the head is broken, increasing the difficulty of attacks based on shared model parameters. We evaluate the privacy-preserving ability of \Method in \Cref{sec:privacy}.

\section{Experimental Setup}

We evaluate \Method on various image/text classification tasks. 
For the image classification tasks, we use four famous datasets, including MNIST~\cite{lecun1998gradient}, Cifar10~\cite{krizhevsky2009learning}, Cifar100~\cite{krizhevsky2009learning} and Tiny-ImageNet~\cite{chrabaszcz2017downsampled} (100K images with 200 classes) using a famous 4-layer CNN~\cite{mcmahan2017communication, luo2021no, geiping2020inverting}. 
To evaluate \Method on a larger backbone model than the 4-layer CNN, we also use ResNet-18~\cite{he2016deep} on Tiny-ImageNet. We set the local learning rate $\eta$ = 0.005 for the 4-layer CNN and $\eta$ = 0.1 for ResNet-18. For the text classification tasks, we use the AG News~\cite{zhang2015character} dataset with the fastText~\cite{joulinetal2017bag} and set $\eta$ = 0.1 for fastText with other settings being the same as image classification tasks. 

We simulate the heterogeneous settings in two widely-used scenarios, \ie, the pathological setting~\cite{mcmahan2017communication, pmlrv139shamsian21a} and practical setting~\cite{NEURIPS2020_18df51b9, li2021model}. For the pathological setting, we sample 2/2/10 classes on MNIST/Cifar10/Cifar100 from a total of 10/10/100 classes for each client with disjoint data. Specifically, similar to FedAvg~\cite{mcmahan2017communication}, we separate clients into groups that own unbalanced data with the same labels. Following MOON~\cite{li2021model}, we create the practical setting through the Dirichlet distribution, denoted as $Dir(\beta)$. Specifically, we sample $q_{c, i} \sim Dir(\beta)$ and allocate a $q_{c, i}$ proportion of the samples of class $c$ to client $i$. 
We set $\beta$ = 0.1 for the default practical setting~\cite{NEURIPS2020_18df51b9, NEURIPS2020_564127c0}. Then, we split the data on each client into a training dataset (75\%) and a test dataset (25\%). 

Following FedAvg, we set the local batch size to 10 and the number of local learning epochs to 1. We run all tasks up to 2000 iterations until all methods converge empirically. Based on pFedMe, FedFomo, and FedRoD, we set the total number of clients to 20 and the client joining ratio $\rho$ = 1 by default. 
Following pFedMe, we report the test accuracy of the best global model for traditional FL methods and the average test accuracy of the best personalized models for pFL methods. We run all the experiments five times and report the mean and standard deviation. Besides, we run all experiments on a machine with two Intel Xeon Gold 6140 CPUs (36 cores), 128G memory, eight NVIDIA 2080 Ti GPUs, and CentOS 7.8. For more results and details, please refer to the Appendix. 

\begin{figure}[!h]
	\centering
	\includegraphics[width=\linewidth]{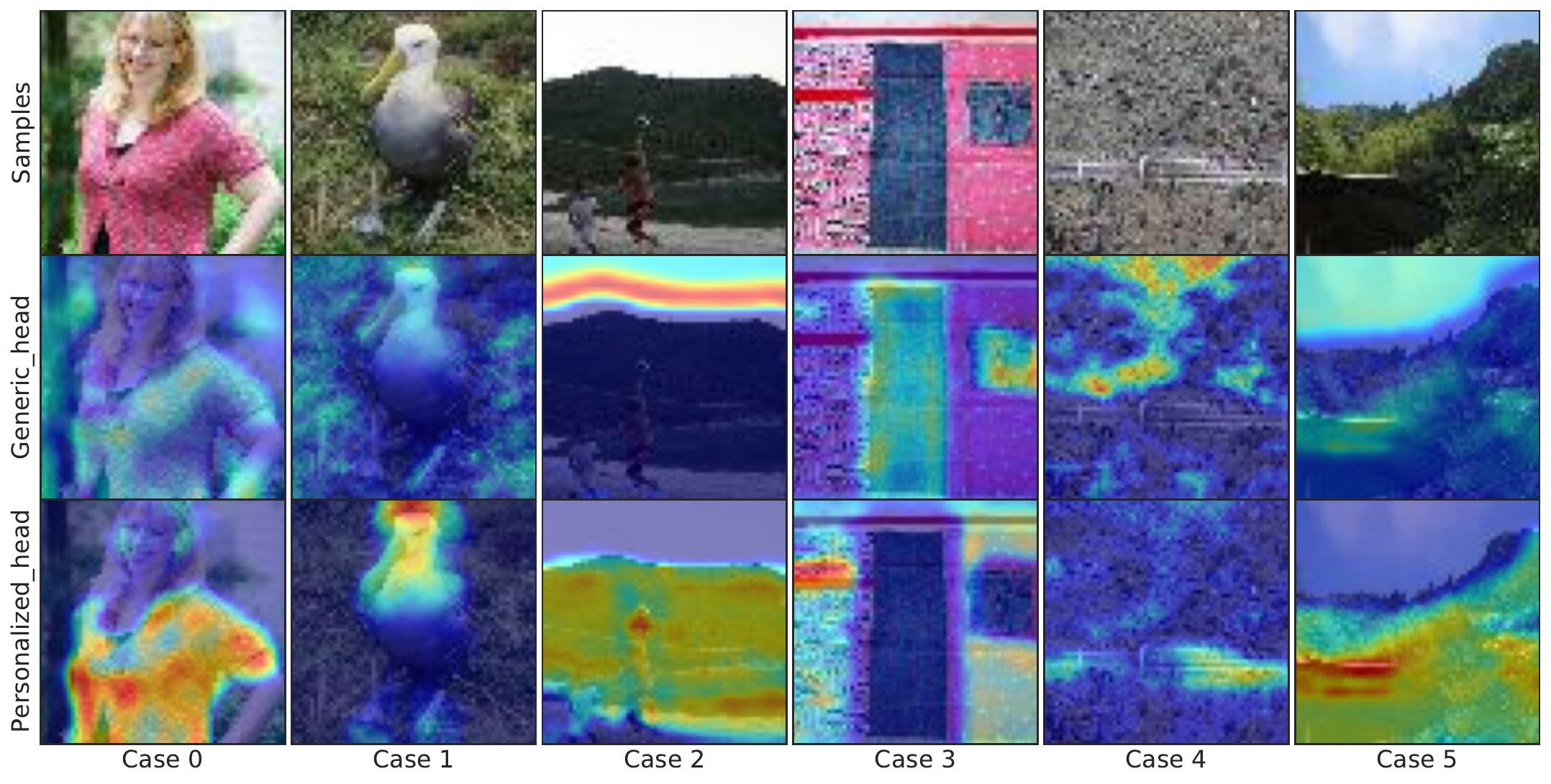}
	\caption{The first row shows six samples from Tiny-ImageNet. The second and third rows respectively show the Grad-CAM visualizations of the learned personalized model with only the global head or the personalized head activated. Highlighted areas are the parts the model pays attention to. }
	\label{fig:cam}
\end{figure}

\begin{figure*}[!ht]
	\centering
	\hfill
	\subfigure[\textit{w.o. GFM}]{\includegraphics[width=0.2\linewidth]{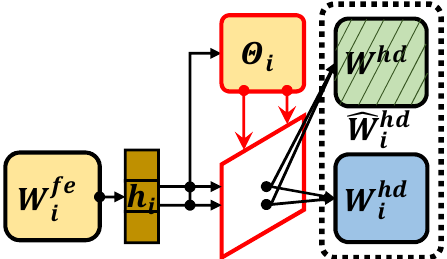}\label{fig:abl_2}}
	\hfill
	\subfigure[\textit{w.o. \Policy}]{\includegraphics[width=0.14\linewidth]{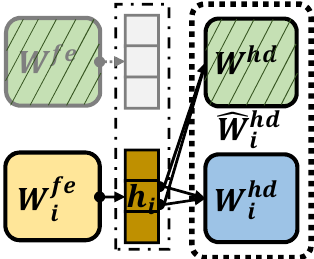}\label{fig:abl_3}}
	\hfill
	\subfigure[\textit{w.o. \Policy \& GFM}]{\includegraphics[width=0.14\linewidth]{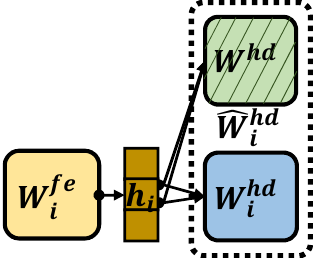}\label{fig:abl_4}}
	\hfill
	\subfigure[\textit{w.o. \Policy \& GH}]{\includegraphics[width=0.14\linewidth]{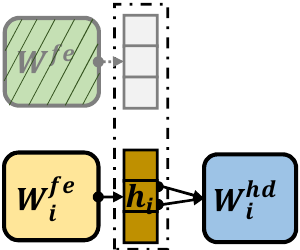}\label{fig:abl_5}}
	\hfill
	\subfigure[\textit{w.o. \Policy \& GFM \& GH}]{\includegraphics[width=0.14\linewidth]{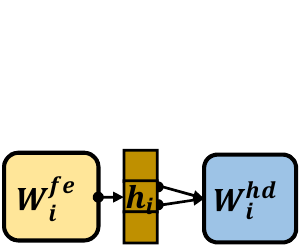}\label{fig:abl_6}}
	\hfill
	\hfill
	\caption{Illustration of variants for module ablation study.}
	\label{fig:abl}
\end{figure*}

\begin{table*}[h]
  \centering
  \caption{The accuracy (\%) on Tiny-ImageNet using ResNet-18 for ablation study.}
  \resizebox{\linewidth}{!}{
    \begin{tabular}{c|cccccccc}
    \toprule 
     \Method & \textit{w.o. cs} & \textit{w.o. ss} & \textit{w.o. cs \& ss} & \textit{w.o. GFM} & \textit{w.o. \Policy} & \textit{w.o. \Policy \& GFM} & \textit{w.o. \Policy \& GH} & \textit{w.o. \Policy \& GFM \& GH}\\
    \midrule
    \textbf{44.18$\pm$0.21} & 43.76$\pm$0.39 & 42.73$\pm$0.26 & 42.25$\pm$0.30 & 42.87$\pm$0.36 & 41.17$\pm$0.18 & 40.06$\pm$0.47 & 35.44$\pm$0.78 & 39.04$\pm$0.79\\
    \bottomrule
    \end{tabular}}
  \label{tab:abl_module}
\end{table*}

\section{Ablation Study}
\subsection{Feature Information Visualization}
\label{sec:act}

To visualize the separated global and personalized feature information when using ResNet-18, we adopt the Grad-CAM~\cite{selvaraju2017grad} on the learned personalized model when only the global head or the personalized head is activated. Six cases from Tiny-ImageNet are shown in \Cref{fig:cam}. 

According to \Cref{fig:cam}, with only the global head activated, the personalized model focuses on relatively global information, such as trees (Case 0 and Case 4), grasses (Case 1), or sky (Case 2 and Case 5) in the background. When we only activate the personalized head, the personalized model focuses on the relatively personalized information, such as foreground (Case 2 and Case 5) or objects (Case 0, Case 1, and Case 4). As for Case 3, the rarely-used pink color is more personalized than the widely-used blue color.

\subsection{Effectiveness of \Policy input} 
To show the effectiveness of each part of the \Policy input, we remove them one by one and obtain the variants: without client-specific vector (\textit{w.o. cs}), without sample-specific vector (\textit{w.o. ss}), without client-specific and sample-specific vector (\textit{w.o. cs \& ss}). For \textit{w.o. cs \& ss}, we regard the randomly initialized frozen vector as the \Policy input, which has the same shape as the sample-specific vector. 

In \Cref{tab:abl_module}, removing either the client-specific vector or the sample-specific vector causes an accuracy decrease. However, \textit{w.o. cs} performs better than \textit{w.o. ss}, so the sample-specific vector is more significant than the client-specific one. According to \Cref{tab:abl_module} and \Cref{tab:pathological}, removing these two kinds of information and using the random vector, \textit{w.o. cs \& ss} still achieves higher accuracy than all the baselines because \Policy module can still learn to separate feature information through the end-to-end training. 

\subsection{Effectiveness of \Method modules} 
To show the effectiveness of each module in \Method, we remove them one by one and obtain the variants: without the frozen global feature extractor and the MMD loss (without GFM for short, \ie, \textit{w.o. GFM}), without \Policy (\textit{w.o. \Policy}), without \Policy and GFM (\textit{w.o. \Policy \& GFM}), without \Policy and the frozen global head (\textit{w.o. \Policy \& GH}), without \Policy, GFM, and the frozen global head (\textit{w.o. \Policy \& GFM \& GH}, similar to FedPer), as shown in \Cref{fig:abl}. It is invalid to keep \Policy while removing the frozen global head since they are a \textit{\textbf{union}} for our feature separating goal. 

In \Cref{tab:abl_module}, without the GFM to align the features, the accuracy of \textit{w.o. GFM} decreases by 1.31\% compared to \Method, but it still outperforms other baselines (see \Cref{tab:pathological}). Without \Policy, the accuracy of \textit{w.o. \Policy} decreases by 3.01\%, so \Policy is more critical than the GFM when the frozen global head exists. Removing both the \Policy and the GFM (\textit{w.o. \Policy \& GFM}) degenerates further than removing one of them, which means that these two modules can facilitate each other. The \Policy and the frozen global head are the key modules in \Method. Without them, the performance of \textit{w.o. \Policy \& GH} degenerates significantly, with a 8.74\% drop compared to \Method. Furthermore, \textit{w.o. \Policy \& GFM \& GH} (removing all the modules) performs better than \textit{w.o. \Policy \& GH}. It means simply adding the GFM to \textit{w.o. \Policy \& GFM \& GH} causes performance degeneration.

\begin{table*}[htbp]
    \centering
    \caption{The accuracy (\%) of the image/text classification tasks in the main experiments.}
    \resizebox{\linewidth}{!}{
      \begin{tabular}{l|ccc|cccccc}
      \toprule
      Settings & \multicolumn{3}{c|}{Pathological setting} & \multicolumn{6}{c}{Default practical setting ($\beta$ = 0.1)} \\
      \midrule
       & MNIST & Cifar10 & Cifar100 & MNIST & Cifar10 & Cifar100 & TINY & TINY* & AG News\\
      \midrule
      FedAvg~\cite{mcmahan2017communication} & 97.93$\pm$0.05 & 55.09$\pm$0.83 & 25.98$\pm$0.13 & 98.81$\pm$0.01 & 59.16$\pm$0.47 & 31.89$\pm$0.47 & 19.46$\pm$0.20 & 19.45$\pm$0.13 & 79.57$\pm$0.17 \\
      FedProx~\cite{MLSYS2020_38af8613} & 98.01$\pm$0.09 & 55.06$\pm$0.75 & 25.94$\pm$0.16 & 98.82$\pm$0.01 & 59.21$\pm$0.40 & 31.99$\pm$0.41 & 19.37$\pm$0.22 & 19.27$\pm$0.23 & 79.35$\pm$0.23 \\
      \midrule
      Per-FedAvg~\cite{NEURIPS2020_24389bfe} & 99.63$\pm$0.02 & 89.63$\pm$0.23 & 56.80$\pm$0.26 & 98.90$\pm$0.05 & 87.74$\pm$0.19 & 44.28$\pm$0.33 & 25.07$\pm$0.07 & 21.81$\pm$0.54 & 93.27$\pm$0.25 \\
      pFedMe~\cite{t2020personalized} & 99.75$\pm$0.02 & 90.11$\pm$0.10 & 58.20$\pm$0.14 & 99.52$\pm$0.02 & 88.09$\pm$0.32 & 47.34$\pm$0.46 & 26.93$\pm$0.19 & 33.44$\pm$0.33 & 91.41$\pm$0.22 \\
      FedAMP~\cite{huang2021personalized} & 99.76$\pm$0.02 & 90.79$\pm$0.16 & 64.34$\pm$0.37 & 99.47$\pm$0.02 & 88.70$\pm$0.18 & 47.69$\pm$0.49 & 27.99$\pm$0.11 & 29.11$\pm$0.15 & 94.18$\pm$0.09 \\
      Ditto~\cite{li2021ditto} & 99.81$\pm$0.00 & 92.39$\pm$0.06 & 67.23$\pm$0.07 & 99.64$\pm$0.00 & 90.59$\pm$0.01 & 52.87$\pm$0.64 & 32.15$\pm$0.04 & 35.92$\pm$0.43 & 95.45$\pm$0.17 \\
      FedPer~\cite{arivazhagan2019federated} & 99.70$\pm$0.02 & 91.15$\pm$0.21 & 63.53$\pm$0.21 & 99.47$\pm$0.04 & 89.22$\pm$0.33 & 49.63$\pm$0.54 & 33.84$\pm$0.34 & 38.45$\pm$0.85 & 95.54$\pm$0.32 \\
      FedRep~\cite{collins2021exploiting} & 99.77$\pm$0.03 & 91.93$\pm$0.14 & 67.56$\pm$0.31 & 99.48$\pm$0.02 & 90.40$\pm$0.24 & 52.39$\pm$0.35 & 37.27$\pm$0.20 & 39.95$\pm$0.61 & 96.28$\pm$0.14 \\
      FedRoD~\cite{chen2021bridging} & 99.90$\pm$0.00 & 91.98$\pm$0.03 & 62.30$\pm$0.02 & 99.66$\pm$0.00 & 89.93$\pm$0.01 & 50.94$\pm$0.11 & 36.43$\pm$0.05 & 37.99$\pm$0.26 & 95.99$\pm$0.08 \\
      FedFomo~\cite{zhang2020personalized} & 99.83$\pm$0.00 & 91.85$\pm$0.02 & 62.49$\pm$0.22 & 99.33$\pm$0.04 & 88.06$\pm$0.02 & 45.39$\pm$0.45 & 26.33$\pm$0.22 & 26.84$\pm$0.11 & 95.84$\pm$0.15 \\
      FedPHP~\cite{li2021fedphp} & 99.73$\pm$0.00 & 90.01$\pm$0.00 & 63.09$\pm$0.04 & 99.58$\pm$0.00 & 88.92$\pm$0.02 & 50.52$\pm$0.16 & 35.69$\pm$3.26 & 29.90$\pm$0.51 & 94.38$\pm$0.12 \\
      \midrule
      \Method & {\bf 99.91$\pm$0.01} & {\bf 92.67$\pm$0.09} & {\bf 71.80$\pm$0.16} & {\bf 99.71$\pm$0.00} & {\bf 91.30$\pm$0.17} & {\bf 59.56$\pm$0.08} & {\bf 43.49$\pm$0.04} & {\bf 44.18$\pm$0.21} & {\bf 96.78$\pm$0.09} \\
      \bottomrule
      \end{tabular}}
    \label{tab:pathological}
  \end{table*}

\section{Evaluation and Analysis}

\subsection{Main Experiments}
\label{sec:main_exp}
Due to the limited space, we use the ``TINY'' and ``TINY*'' to represent using the 4-layer CNN on Tiny-ImageNet and using ResNet-18 on Tiny-ImageNet, respectively. \Cref{tab:pathological} shows that \Method outperforms all the baselines when using either the 4-layer CNN or the ResNet-18, especially on relatively challenging tasks. In the default practical setting on Cifar100, \Method exceeds the best baseline (Ditto) by \textbf{6.69\%}. Our \Policy only introduces an additional 0.527M (million) parameters on each client, which is 9.25\% and 4.67\% of the parameters in the 4-layer CNN (5.695M) and the ResNet-18 (11.279M), respectively. In the following, we analyze \emph{\textbf{why}} \Method outperforms all the baselines. 

In \Cref{tab:pathological}, FedAvg and FedProx perform poorly, as the global model cannot fit the local data well on all the clients. They directly feed features to the global head, regardless of the personalized information in the features. In contrast, \Method separates and feeds the global information and the personalized information in the features to the global head and the personalized head, respectively. 

Per-FedAvg performs poorly among pFL methods, as the aggregated learning trend can hardly meet the trend of each personalized model. In contrast, \Method considers personalization in a sample-specific manner conditioned by the client-specific vector, which meets the demand of each client, thus performing better. 

pFedMe and FedAMP utilize regularization terms to extract information from the local model and the client-specific server model, respectively. However, excessively concentrating on personalization is not beneficial to the collaborative goal of FL. Since Ditto extracts global information from the global model, it performs better than pFedMe and FedAMP. Like Ditto, \Method also takes advantage of global information for each client. 

FedPer and FedRep only share the feature extractor without sharing heads. They ignore some global information in the head part, so they perform worse than \Method. FedRoD bridges the goal of traditional FL and pFL by learning two heads with two objectives. However, these two goals are competing~\cite{chen2021bridging}, so FedRoD performs worse than FedRep, which also learns a personalized head but only focuses on the goal of pFL. 
Like FedRep,
\Method only focuses on the pFL goal, thus performing the best. 

Similar to FedAMP, FedFomo aggregates client models with client-specific weights, thus losing some global information. FedPHP transfers the global information only in the global feature extractor through the MMD loss. Although it achieves excellent performance, FedPHP loses the global information in the global head during local training, so it performs worse than \Method.

\subsection{Computing and Communication Overhead}

Here, we focus on the training phase. We report the total time and the number of iterations required for each method to converge and calculate the average time consumption in each iteration, as shown in \Cref{tab:app}. Ditto and pFedMe cost more time in each iteration than most methods since the additional personalized model training takes much extra time. 
Compared to most baselines, \eg, Per-FedAvg, pFedMe, Ditto, FedRep, and FedPHP, \Method costs less training time in each iteration. In \Method, the parameters in the \Policy module only require an additional 4.67\% communication overhead per iteration when using ResNet-18 compared to FedAvg. 

\begin{table}[h]
  \centering
  \caption{The computing time and communication iterations on Tiny-ImageNet using ResNet-18.}
  \label{tab:nlp}
  \resizebox{!}{!}{
    \begin{tabular}{l|rrr}
    \toprule
    & Total time & Iterations & Avg. time\\
    \midrule
    FedAvg & 365 min & 230 & 1.59 min\\
    FedProx & 325 min & 163 & 1.99 min\\
    \midrule
    Per-FedAvg & 121 min & 34 & 3.56 min\\
    pFedMe & 1157 min & 113 & 10.24 min\\
    FedAMP & 92 min & 60 & 1.53 min\\
    Ditto & 318 min & 27 & 11.78 min\\
    FedPer & 83 min & 43 &  1.92 min\\
    FedRep & 471 min & 115 & 4.09 min\\
    FedRoD & 87 min & 50 & 1.74 min\\
    FedFomo & 193 min & 71 & 2.72 min\\
    FedPHP & 264 min & 65 & 4.06 min\\
    \midrule
    \Method & 204 min & 74 & 2.75 min\\
    \bottomrule
    \end{tabular}}
    \label{tab:app}
\end{table}

\begin{table*}[ht]
  \centering
  \caption{The accuracy (\%) of the image/text classification tasks for heterogeneity and scalability.}
  \resizebox{\linewidth}{!}{
    \begin{tabular}{l|ccc|cccccc}
    \toprule
    & \multicolumn{3}{c|}{Heterogeneity} & \multicolumn{6}{c}{Scalability}\\
    \midrule
    Datasets & \multicolumn{2}{c}{TINY} & AG News & \multicolumn{6}{c}{Cifar100}\\
    \midrule
     & $\beta$ = 0.01 & $\beta$ = 0.5 & $\beta$ = 1 & $N$ = 10 & $N$ = 30 & $N$ = 50 & $N$ = 100 & $N$ = 200 & $N$ = 500 \\
    \midrule
    FedAvg & 15.70$\pm$0.46 & 21.14$\pm$0.47 & 87.12$\pm$0.19 & 31.47$\pm$0.01 & 31.15$\pm$0.05 & 31.90$\pm$0.27 & 31.95$\pm$0.37 & 31.20$\pm$0.58 & 29.51$\pm$0.73 \\
    FedProx & 15.66$\pm$0.36 & 21.22$\pm$0.47 & 87.21$\pm$0.13 & 31.24$\pm$0.08 & 31.21$\pm$0.08 & 31.94$\pm$0.30 & 31.97$\pm$0.24 & 31.22$\pm$0.62 & 29.84$\pm$0.81 \\
    \midrule
    Per-FedAvg & 39.39$\pm$0.30 & 16.36$\pm$0.13 & 87.08$\pm$0.26 & 37.24$\pm$0.12 & 41.57$\pm$0.21 & 44.31$\pm$0.20 & 36.07$\pm$0.24 & --- & --- \\
    pFedMe & 41.45$\pm$0.14 & 17.48$\pm$0.61 & 87.08$\pm$0.18 & 44.06$\pm$0.29 & 47.04$\pm$0.28 & 48.36$\pm$0.64 & 46.45$\pm$0.18 & 39.55$\pm$0.61 & 31.30$\pm$0.89 \\
    FedAMP & 48.42$\pm$0.06 & 12.48$\pm$0.21 & 83.35$\pm$0.05 & 49.23$\pm$0.18 & 45.33$\pm$0.04 & 44.39$\pm$0.35 & 40.43$\pm$0.17 & 35.40$\pm$0.70 & \textit{diverged} \\
    Ditto & 50.62$\pm$0.02 & 18.98$\pm$0.05 & 91.89$\pm$0.17 & 52.32$\pm$0.19 & 52.53$\pm$0.42 & 54.22$\pm$0.04 & 52.89$\pm$0.22 & 35.18$\pm$0.53 & 30.24$\pm$0.72 \\
    FedPer & 51.83$\pm$0.22 & 17.31$\pm$0.19 & 91.85$\pm$0.24 & 50.31$\pm$0.19 & 44.98$\pm$0.20 & 44.22$\pm$0.18 & 40.37$\pm$0.41 & 34.99$\pm$0.48 & 30.56$\pm$0.59 \\
    FedRep & 55.43$\pm$0.15 & 16.74$\pm$0.09 & 92.25$\pm$0.20 & 52.89$\pm$0.10 & 50.24$\pm$0.01 & 47.41$\pm$0.18 & 44.61$\pm$0.20 & 36.79$\pm$0.60 & 31.92$\pm$0.71 \\
    FedRoD & 49.17$\pm$0.06 & 23.23$\pm$0.11 & 92.16$\pm$0.12 & 49.83$\pm$0.07 & 50.11$\pm$0.03 & 49.38$\pm$0.01 & 46.65$\pm$0.22 & 43.53$\pm$0.86 & 34.61$\pm$0.98 \\
    FedFomo & 46.36$\pm$0.54 & 11.59$\pm$0.11 & 91.20$\pm$0.18 & 46.71$\pm$0.23 & 43.20$\pm$0.05 & 42.56$\pm$0.33 & 38.91$\pm$0.08 & 34.79$\pm$0.71 & 29.24$\pm$1.28 \\
    FedPHP & 48.63$\pm$0.02 & 21.09$\pm$0.07 & 90.52$\pm$0.19 & 49.32$\pm$0.19 & 49.28$\pm$0.06 & 52.44$\pm$0.16 & 49.70$\pm$0.31 & 34.48$\pm$0.33 & 30.26$\pm$0.84 \\
    \midrule
    \Method & {\bf 56.31$\pm$0.39} & {\bf 27.66$\pm$0.16} & {\bf 92.89$\pm$0.10} & {\bf 58.36$\pm$0.02} & {\bf 56.93$\pm$0.19} & {\bf 55.43$\pm$0.21} & {\bf 53.81$\pm$0.32} & {\bf 44.86$\pm$0.87} & {\bf 35.87$\pm$0.52} \\
    \bottomrule
    \end{tabular}}
  \label{tab:beta}
\end{table*}

\subsection{Different Heterogeneity Degrees}
\label{sec:hete}

In addition to \Cref{tab:pathological}, we conduct experiments on the settings with different degrees of heterogeneity on Tiny-ImageNet and AG News by varying $\beta$. The smaller the $\beta$ is, the more heterogeneous the setting is. We show the accuracy in \Cref{tab:beta}, where \Method still outperforms the baselines. Most pFL methods achieve higher accuracy than traditional FL methods in the more heterogeneous setting. In the setting with a larger $\beta$, most of them cannot achieve higher accuracy than FedAvg on Tiny-ImageNet. In contrast, the methods that utilize global information during local learning (FedPHP, FedRoD, and \Method) maintain excellent performance. 
FedRoD performs worse than FedRep, as the latter focuses only on the goal of pFL. 
pFedMe and FedAMP perform poorly among pFL methods. Their accuracy is lower than traditional FL methods when $\beta$ = 1.

\subsection{Scalability with Different Client Amounts}

Following MOON~\cite{li2021model}, we conduct another six experiments (\ie, $N$ = 10, $N$ = 30, $N$ = 50, $N$ = 100, $N$ = 200, and $N$ = 500) to study the scalability of \Method and keep other settings unchanged. Per-FedAvg requires more data than other methods, as meta-learning requires at least two batches of data, which is invalid on some clients in our unbalanced settings when $N$ $\ge$ 200. Since the total data amount is constant on Cifar100, the local data amount (on average) decreases as the client amount increases. With both $N$ and local data amount changing, it is unreasonable to compare the results among different $N$ in \Cref{tab:beta}. 
Some pFL methods, including Per-FedAvg and pFedMe, achieve relatively poor performance in the setting with $N$ = 10, where few clients (\eg, hospitals) participate in FL, and each of them possesses a large data repository. 
When $N$ = 500 (\eg, mobile smartphones), each client only has 90 samples for training on average, which is not enough for the weight calculation in FedFomo, so it performs worse than FedAvg. FedAMP diverges as it is hard to find similar clients when they have little data. According to \Cref{tab:beta}, \Method still outperforms all the baselines.

\begin{table}[ht]
  \centering
  \caption{The accuracy (\%) on Cifar100 for scalability.}
  \resizebox{!}{!}{
    \begin{tabular}{l|ccc}
    \toprule
    & $N$ = 10|50 & $N$ = 30|50 & $N$ = 50 \\
    \midrule
    FedAvg & 25.28$\pm$0.32 & 29.04$\pm$0.21 & 31.90$\pm$0.27 \\
    FedProx & 25.65$\pm$0.34 & 29.04$\pm$0.36 & 31.94$\pm$0.30 \\
    \midrule
    Per-FedAvg & 40.20$\pm$0.21 & 42.96$\pm$0.42 & 44.31$\pm$0.20 \\
    pFedMe & 40.27$\pm$0.54 & 42.19$\pm$0.38 & 48.36$\pm$0.64 \\
    FedAMP & 43.57$\pm$0.30 & 43.18$\pm$0.31 & 44.39$\pm$0.35 \\
    Ditto & 48.23$\pm$0.35 & 50.98$\pm$0.29 & 54.22$\pm$0.04 \\
    FedPer & 43.64$\pm$0.42 & 43.54$\pm$0.43 & 44.22$\pm$0.18 \\
    FedRep & 46.85$\pm$0.12 & 47.63$\pm$0.26 & 47.41$\pm$0.18 \\
    FedRoD & 46.32$\pm$0.02 & 49.15$\pm$0.12 & 49.38$\pm$0.01 \\
    FedFomo & 41.53$\pm$0.45 & 40.69$\pm$0.41 & 42.56$\pm$0.33 \\
    FedPHP & 45.71$\pm$0.21 & 48.65$\pm$0.24 & 52.44$\pm$0.16 \\
    \midrule
    \Method & {\bf 50.93$\pm$0.34} & {\bf 54.31$\pm$0.25} & {\bf 55.43$\pm$0.21} \\
    \bottomrule
    \end{tabular}}
  \label{tab:scalability*}
\end{table}

To simulate a real-world scenario where more clients means more total data amount in FL, we consider the setting Cifar100 ($\beta$ = 0.1, $\rho$ = 1, and $N$ = 50) used above as the base setting and randomly sample 10 and 30 clients from existing 50 clients to form the Cifar100 ($\beta$ = 0.1, $\rho$ = 1, and $N$ = 10|50) and Cifar100 ($\beta$ = 0.1, $\rho$ = 1, and $N$ = 30|50) settings, respectively. When we increase the client amount, the accuracy increases as more data are utilized to train the globally shared modules, which facilitates information transfer among clients. The superior performance of \Method in \Cref{tab:scalability*} shows its scalability in this real-world scenario. 

\subsection{Large Local Epochs}

\begin{table}[ht]
  \centering
  \caption{The accuracy (\%) on Cifar10 in the default practical setting with large local epochs.}
  \resizebox{\linewidth}{!}{
    \begin{tabular}{l|cccc}
    \toprule
    Local epochs & 5 & 10 & 20 & 40\\
    \midrule
    FedAvg & 57.51$\pm$0.35 & 57.55$\pm$0.32 & 57.28$\pm$0.23 & 56.27$\pm$0.29\\
    FedProx & 57.48$\pm$0.28 & 57.69$\pm$0.31 & 57.53$\pm$0.33 & 56.18$\pm$0.24\\
    \midrule
    Per-FedAvg & 86.13$\pm$0.12 & 86.09$\pm$0.19 & 85.57$\pm$0.15 & 85.45$\pm$0.16\\
    pFedMe & 88.72$\pm$0.02 & 88.58$\pm$0.17 & 88.37$\pm$0.14 & 88.16$\pm$0.20\\
    FedAMP & 88.72$\pm$0.21 & 88.77$\pm$0.27 & 88.76$\pm$0.30 & 88.70$\pm$0.26\\
    Ditto & 90.79$\pm$0.21 & 90.59$\pm$0.06 & 90.34$\pm$0.23 & 90.02$\pm$0.38\\
    FedPer & 89.62$\pm$0.12 & 89.73$\pm$0.31 & 89.79$\pm$0.35 & 89.49$\pm$0.55\\
    FedRep & 90.20$\pm$0.41 & 90.08$\pm$0.26 & 89.46$\pm$0.13 & 89.22$\pm$0.25\\
    FedRoD & 89.71$\pm$0.32 & 89.11$\pm$0.33 & 88.13$\pm$0.21 & 87.55$\pm$0.28\\
    FedFomo & 88.39$\pm$0.15 & 88.43$\pm$0.16 & 88.41$\pm$0.13 & 88.13$\pm$0.32\\
    FedPHP & 90.29$\pm$0.37 & 90.03$\pm$0.23 & 89.92$\pm$0.27 & 89.87$\pm$0.26\\
    \midrule
    \Method & {\bf 91.13$\pm$0.34} & {\bf 91.24$\pm$0.31} & {\bf 91.02$\pm$0.28} & {\bf 90.86$\pm$0.37}\\
    \bottomrule
    \end{tabular}}
    \label{tab:largeE}
\end{table}

Large local epochs can reduce total communication iterations but increase computing overhead per iteration for most of the methods in FL~\cite{mcmahan2017communication}. 
With larger local epochs, \Method can still maintain its superiority as shown in \Cref{tab:largeE}. Most of the methods perform worse with larger local epochs since more local training aggravates the discrepancy among client models, which is adverse to server aggregation. For example, the accuracy of FedRoD drops by 2.16\% when the number of local epochs increases from 5 to 40.

\subsection{Clients Accidentally Dropping Out}
\label{sec:adapt}

\begin{table}[ht]
  \centering
  \caption{The accuracy (\%) on Cifar100 ($N$ = 50, $\beta$ = 0.1) when clients accidentally drop out.}
  \resizebox{!}{!}{
    \begin{tabular}{l|ccc}
    \toprule
    & $\rho$ = 1 & $\rho \in [0.5, 1]$ & $\rho \in [0.1, 1]$ \\
    \midrule
    Per-FedAvg & 44.31$\pm$0.20 & 43.66$\pm$1.38 & 43.63$\pm$1.07\\
    pFedMe & 48.36$\pm$0.64 & 43.28$\pm$0.85 & 41.71$\pm$1.02\\
    FedAMP & 44.39$\pm$0.35 & 42.91$\pm$0.08 & 42.92$\pm$0.14\\
    Ditto & 50.59$\pm$0.22 & 49.78$\pm$0.36 & 48.33$\pm$3.27\\
    FedPer & 44.22$\pm$0.18 & 44.12$\pm$0.21 & 44.07$\pm$0.27\\
    FedRep & 47.41$\pm$0.18 & 46.93$\pm$0.21 & 46.61$\pm$0.22\\
    FedRoD & 49.38$\pm$0.01 & 49.07$\pm$0.43 & 47.80$\pm$1.35\\
    FedFomo & 42.56$\pm$0.33 & 40.96$\pm$0.02 & 40.93$\pm$0.07\\
    FedPHP & 50.23$\pm$0.12 & 45.19$\pm$0.07 & 44.43$\pm$0.12\\
    \midrule
    \Method & {\bf 54.81$\pm$0.20} & {\bf 54.68$\pm$0.35} & {\bf 54.20$\pm$0.21}\\
    \bottomrule
    \end{tabular}}
  \label{tab:drop}
\end{table}

Due to the changing network connection quality, some clients may accidentally (randomly) drop out at one iteration and become active again at another iteration, which frequently happens in the mobile settings. We compare the performance of pFL methods when some clients accidentally drop out, as shown in \Cref{tab:drop}. Instead of using the constant $\rho$, we randomly choose a value within a given range for $\rho$ in each iteration. The larger the range of $\rho$ is, the more unstable the setting is. It simulates a more practical setting with a random drop-out rate than the settings used by the SOTA methods, which set a constant drop-out rate in all iterations. 

Most pFL methods suffer from an accuracy decrease in unstable settings. pFedMe and FedPHP have up to 6.65\% and 9.80\% accuracy decrease, respectively, compared to $\rho$ = 1 in \Cref{tab:drop}. Some methods, such as FedRep, and FedRoD, perform worse with a larger range of $\rho$. The standard deviation of Per-FedAvg, pFedMe, Ditto, and FedRoD is greater than 1\% when $\rho \in [0.1, 1]$, which means their performance is unstable with the random $\rho$. Since \Policy separates feature information automatically, \Method can adapt to the changing environments thus still maintaining superiority and stable performance in these unstable settings.

\section{Effect of the Hyper-parameter $\lambda$}
To guide the learned features to fit the frozen global head, we use the hyper-parameter $\lambda$ to control the importance of MMD loss that aligns the outputs of the personalized feature extractor and the outputs of the global feature extractor. The larger the $\lambda$ is, the closer these two outputs are. 

\begin{table}[ht]
  \centering
  \caption{The accuracy (\%) on Tiny-ImageNet using the 4-layer CNN in three practical settings.}
  \resizebox{\linewidth}{!}{
    \begin{tabular}{l|c*{6}{c}}
    \toprule & $\lambda=1$ & $\lambda=2$ & $\lambda=5$ & $\lambda=10$ & $\lambda=50$\\
    \midrule
    $\beta=0.01$ & 56.56$\pm$0.35 & \textbf{56.71$\pm$0.32} & 56.31$\pm$0.39 & 54.48$\pm$0.10 & 9.73$\pm$0.02\\
    $\beta$ = 0.1 & 41.67$\pm$0.17 & 42.75$\pm$0.03 & \textbf{43.49$\pm$0.04} & 42.83$\pm$0.07 & 8.14$\pm$0.06\\
    $\beta=0.5$ & 24.95$\pm$0.15 & 26.55$\pm$0.23 & \textbf{27.66$\pm$0.16} & 26.95$\pm$0.27 & 4.54$\pm$0.04\\
    \bottomrule
    \end{tabular}}
  \label{tab:lam_tiny}
\end{table}

From \Cref{tab:lam_tiny}, the accuracy first increases and then decreases as $\lambda$ increases, which is similar among three settings with different degrees of heterogeneity. By assigning a proper value to $\lambda$, the personalized feature extractor can learn the information from the local data while guiding the output features to fit the frozen global head. When the value of $\lambda$ is overlarge (\eg, $\lambda=50$), the personalized feature extractor can hardly learn from the local data. Instead, it tends to output similarly to the frozen global feature extractor. To pay more attention to the local data in a more heterogeneous setting (\eg, $\beta=0.01$), \Method requires a relatively smaller $\lambda$, as the global information plays a less critical role in this situation. 

\section{Policy Study}
\label{sec:policy}

\begin{figure}[h]
	\centering
	\subfigure[PIR change on client \#0.]{\includegraphics[width=0.39\linewidth]{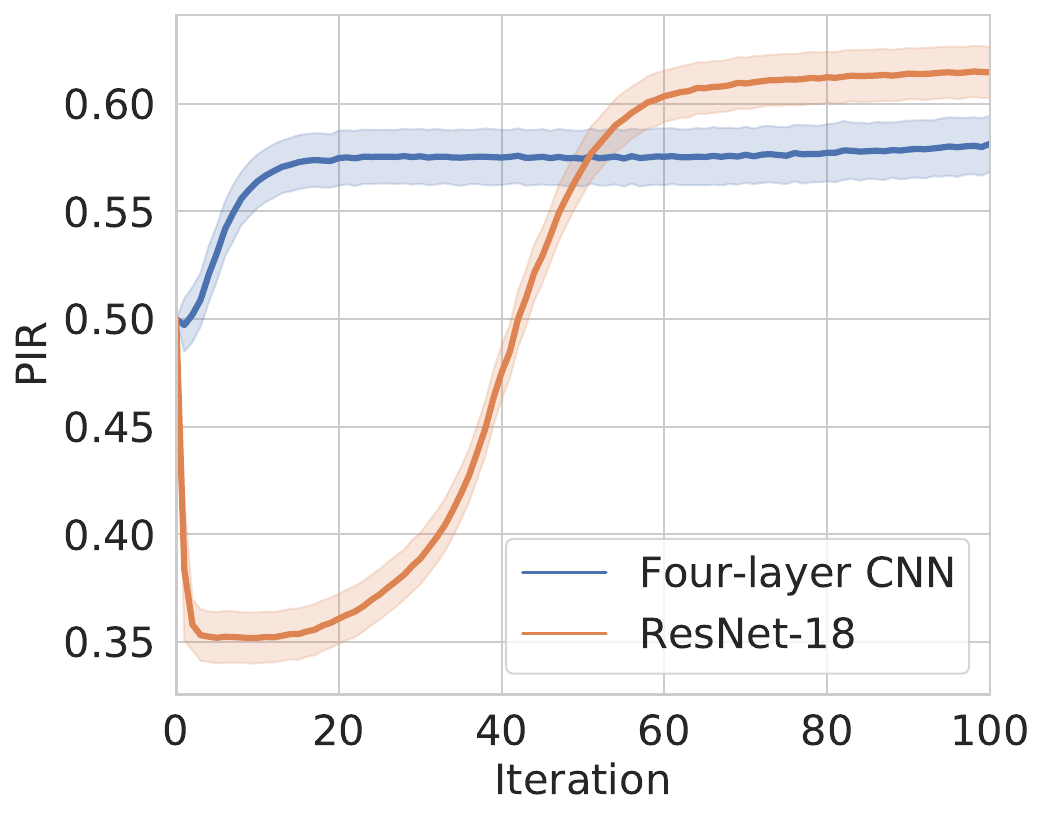}\label{fig:policy_a}}
    \hfill
	\subfigure[${\bm s}_i$ distribution of test samples on all clients.]{\includegraphics[width=0.6\linewidth]{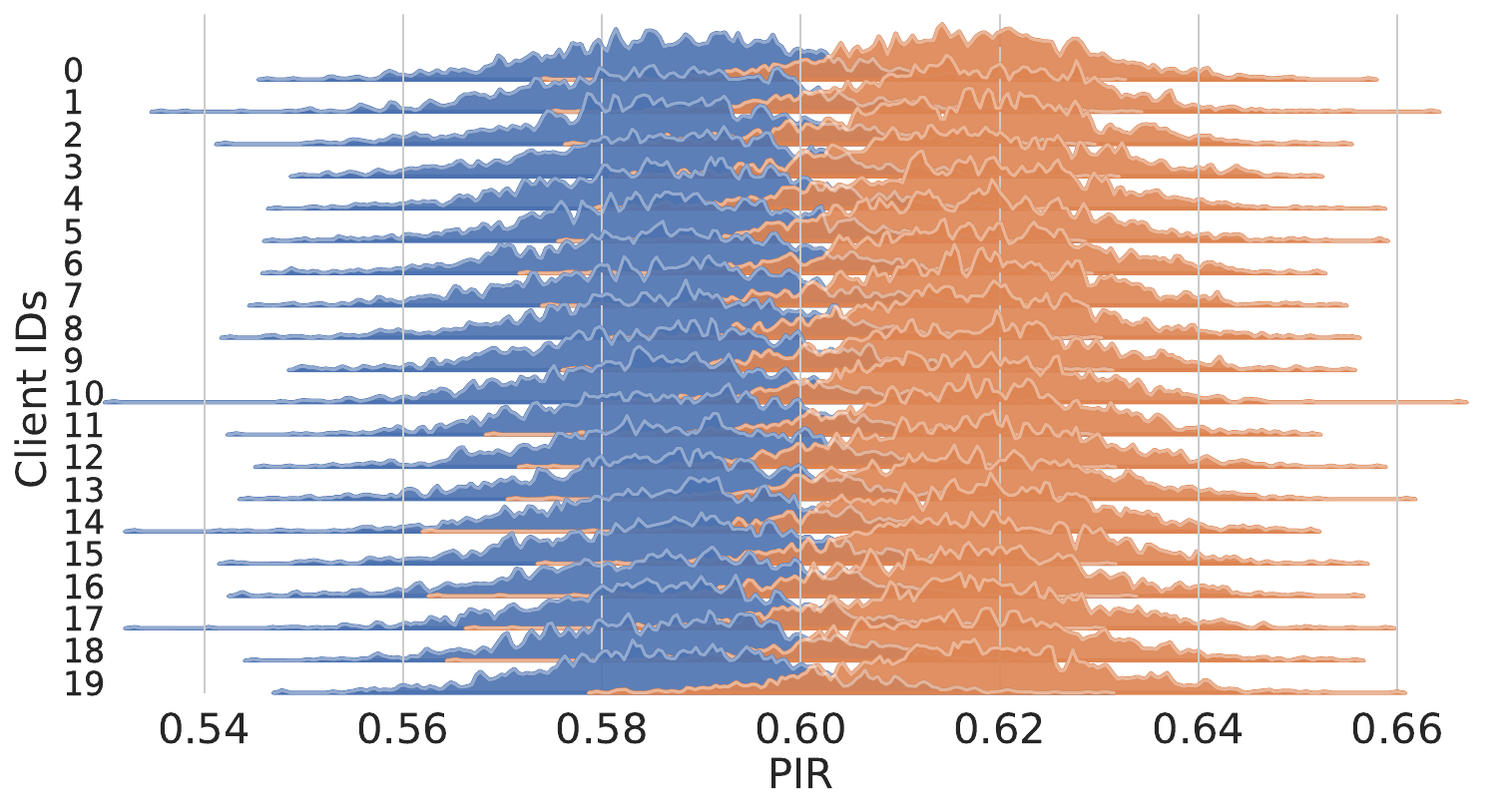}\label{fig:policy_b}}
	\caption{Visualizations for PIR and ${\bm s}_i$ distribution on Tiny-ImageNet in the default practical setting. \textcolor{blue_}{Blue color} and \textcolor{orange_}{orange color} represent the figures for the \textcolor{blue_}{4-layer CNN} and \textcolor{orange_}{ResNet-18}, respectively. We draw PIR change curves for training samples. Best viewed in color. }
	\label{fig:policy}
\end{figure}

We show the policy change for the training samples and the generated policies for all the test samples during inference in \Cref{fig:policy}. For clarity, we collect all the sample-specific ${\bm s}_i$ on each client and average them to obtain $\overline{{\bm s}}_i$. Then we further average the elements in $\overline{{\bm s}}_i$ to generate one scalar, which is called personalization identification ratio (PIR): ${\rm PIR}_i:=\frac{1}{K} \sum^{K}_k \overline{s}^k_i, i\in [N]$, where $\overline{s}^k_i$ is the $k$th element in the policy $\overline{{\bm s}}_i$. 

When using diverse backbones with different feature extraction abilities, the policies vary in both PIR change and ${\bm s}_i$ distribution. As shown in \Cref{fig:policy_a}, on client \#0, PIR increases from the initial value of 0.50 to around 0.58 in the first 20 iterations and remains almost unchanged using the 4-layer CNN. However, when using ResNet-18, PIR decreases first and then increases rapidly to around 0.61, which means that the features extracted by the feature extractor in ResNet-18 contain more global feature information in early iterations, and our \Policy can automatically capture this dynamic characteristic during all FL iterations. In \Cref{fig:policy_b}, the value range of ${\bm s}_i$ varies among clients, as they contain diverse samples. For example, the ${\bm s}_i$ range on client \#10 is the largest among clients. Although the policies are different for the samples, the mean values of ${\bm s}_i$ are similar among clients when using one specific backbone, as shown in \Cref{fig:policy_b}. The values of ${\bm s}_i$ are all larger than 0.5 during inference, which means the learned features contain more personalized feature information than global feature information on clients in these scenarios.

\section{Conclusion}
\label{sec:conc}

We propose a Federated Conditional Policy (\Method) method that generates a policy for each sample to separate its features into the global feature information and the personalized feature information, then processes them by the global head and the personalized head, respectively. \Method outperforms eleven SOTA methods by up to 6.69\% under various settings with excellent privacy-preserving ability. Besides, \Method also maintains excellent performance when some clients accidentally drop out. 

\begin{acks}
    This work was supported in part by the Shanghai Key Laboratory of Scalable Computing and Systems, National Key R\&D Program of China (2022YFB4402102), Internet of Things special subject program, China Institute of IoT (Wuxi), Wuxi IoT Innovation Promotion Center (2022SP-T13-C), Industry-university-research Cooperation Funding Project from the Eighth Research Institute in China Aerospace Science and Technology Corporation (Shanghai) (USCAST2022-17), and Intel Corporation (UFunding 12679). The work of H. Wang was supported in part by the NSF grant CRII-OAC-2153502. Ruhui Ma is the corresponding author. 
\end{acks}


\bibliographystyle{ACM-Reference-Format}
\balance
\bibliography{main}

\clearpage

\appendix

\section{Convergence Analysis}

Recall that our objective is
\begin{equation}
    \{{\bm W}_1, \ldots, {\bm W}_N\} = \argmin \ \mathcal{G}(\mathcal{F}_1, \ldots, \mathcal{F}_N),
\end{equation}
where $\mathcal{F}_i, \forall i \in [N]$ is the local loss and $\mathcal{G}(\mathcal{F}_1, \ldots, \mathcal{F}_N) = \sum^{N}_{i=1} n_i \mathcal{F}_i$. During the training phase, the value of $\mathcal{G}$ is the training loss of \Method. To study the convergence of \Method, we denote the loss calculated with the trained personalized models after local learning as $loss_{aft}$ and the loss calculated with the initialized personalized models before local learning as $loss_{bef}$. Except for the loss values, we also evaluate the corresponding test accuracy, calculated by averaging the accuracy of all the personalized models on the corresponding local test datasets of clients. 

\begin{figure}[h]
	\centering
	\subfigure[Training loss ($\mathcal{G}$) curves]{\includegraphics[width=\linewidth]{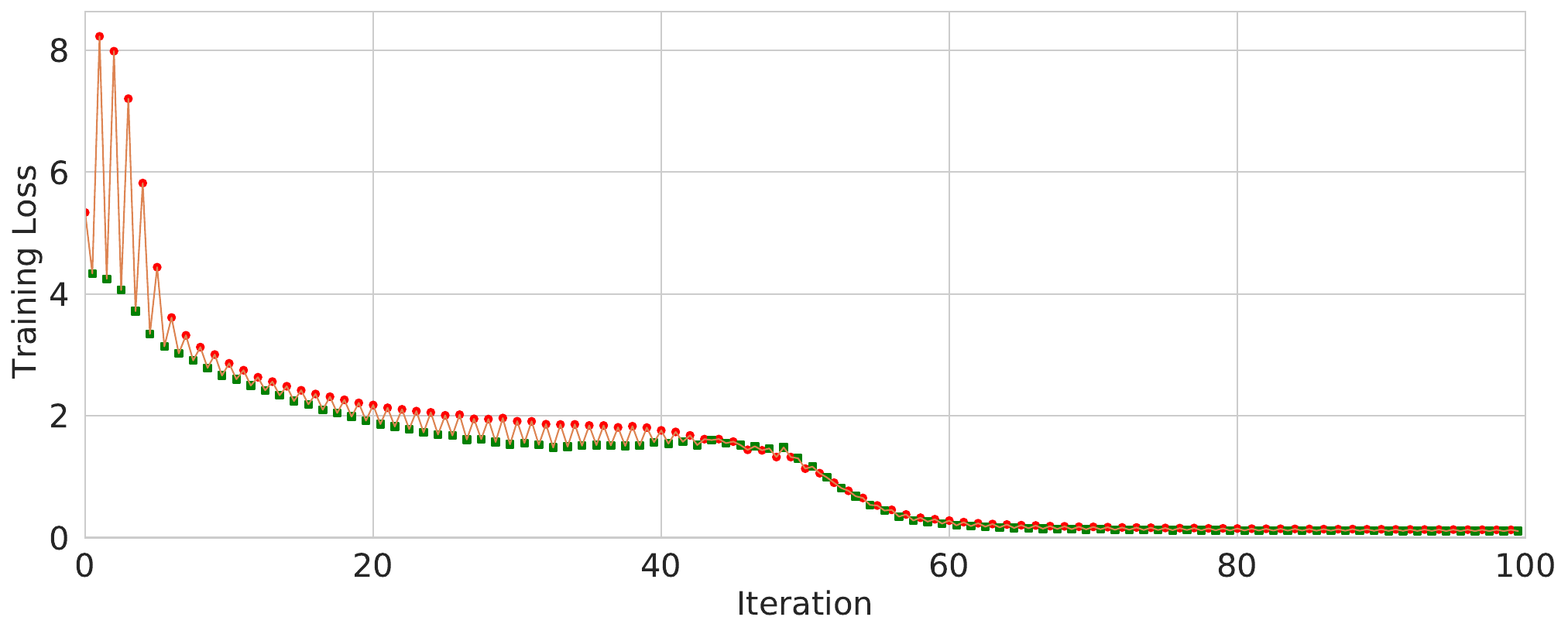}}
	\subfigure[Test accuracy curves]{\includegraphics[width=\linewidth]{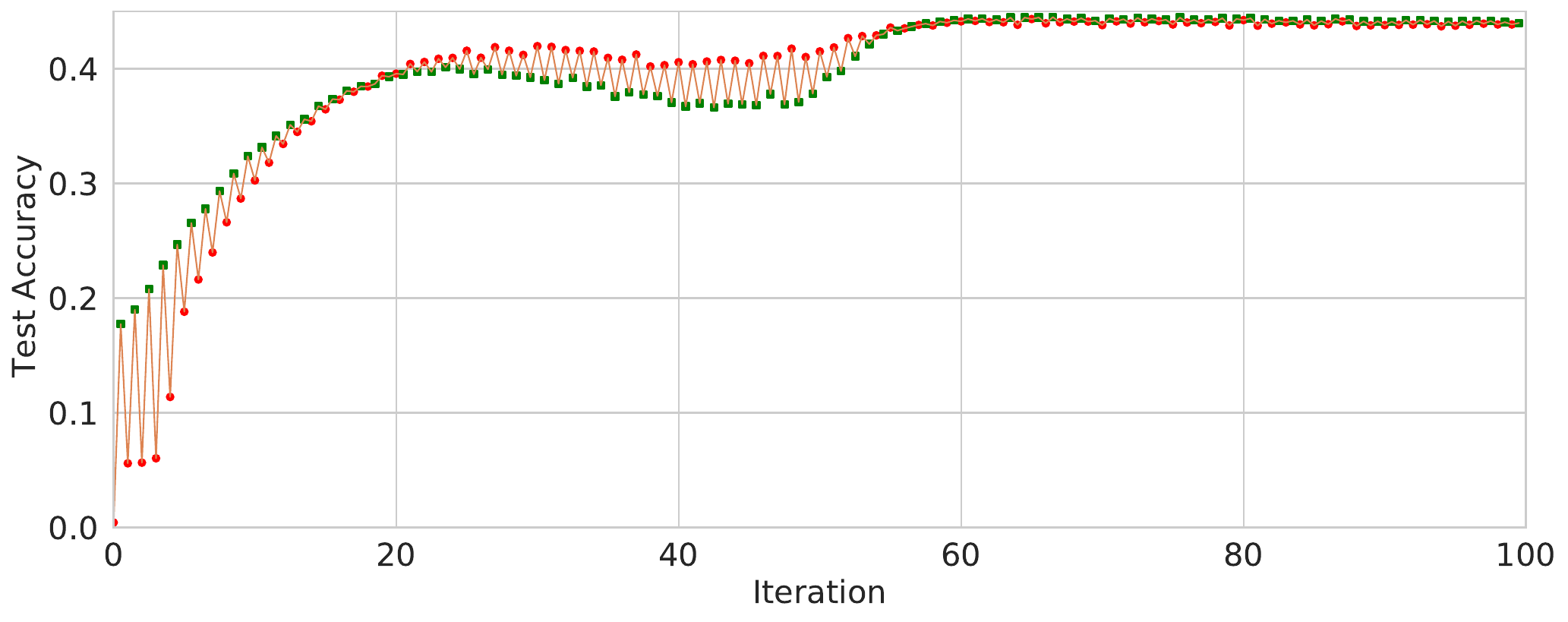}}
	\caption{The training loss curves and test accuracy curves when using ResNet-18 on Tiny-ImageNet in the default practical setting. The \textcolor{red_}{red circles} and \textcolor{green_}{green cubes} represent the results evaluated \textcolor{red_}{before local learning} and \textcolor{green_}{after local learning}, respectively. Best viewed in color. }
	\label{fig:converge}
\end{figure}

To empirically analyze the convergence of \Method, we draw the training loss curves and test accuracy curves for our \Method when using ResNet-18, as shown in \Cref{fig:converge}. On Tiny-ImageNet in the default practical setting, $loss_{aft}$ becomes close to $loss_{bef}$ after 74 iterations, and both of them reach the minimum value meanwhile. In other words, \Method converges after training around 74 iterations. With the training loss decreasing, the test accuracy increases. Both the loss curve and the accuracy curve fluctuate before iteration 56 when using ResNet-18 due to the policy update, as shown in \Cref{fig:policy} in the main body of this paper. 

\section{Privacy-Preserving Ability}
\label{sec:privacy}

Here, following a traditional FL method FedCG~\cite{wufedcg}, we consider a semi-honest scenario where the server follows the FL protocol but may recover original data from a victim client with its model updates via Deep Leakage from Gradients (DLG) attack~\cite{zhu2019deep}. 
Among the baselines in our paper, there are two categories in terms of information transmission between the server and clients. Methods in Category 1 share the parameters in the entire backbone model, such as FedAvg, FedProx, Per-FedAvg, pFedMe, Ditto, FedRoD, FedFomo, and FedPHP. Methods in Category 2 only share the parameters in the feature extractor, such as FedPer and FedRep. Without loss of generality, we select the most famous methods in each category as the representative baselines: FedAvg for Category 1 and FedPer for Category 2. 
Also following FedCG, we provide the experimental results in \Cref{tab:dlg} to evaluate the privacy-preserving ability of \Method with representative baselines in Peak Signal-to-Noise Ratio (PSNR). The lower value of PSNR shows better privacy-preserving ability. The results in \Cref{tab:dlg} show the superiority of \Method.

\begin{table}[h]
  \centering
  \caption{PSNR on Cifar100 in the default practical setting.}
  \resizebox{!}{!}{
    \begin{tabular}{l|cc|c}
    \toprule
     & FedAvg & FedPer & \Method \\
    \midrule
    PSNR (dB, $\downarrow$) & 7.30 & 7.94 & \textbf{6.94} \\
    \bottomrule
    \end{tabular}}
    \label{tab:dlg}
\end{table}

\section{Conditional Policy Network Design}

By default, our \Policy consists of a fully connected (FC) layer~\cite{lecun2015deep} and a layer-normalization layer~\cite{ba2016layer} (LN for short) followed by the ReLU activation function~\cite{li2017convergence}. Here, we investigate how different designs affect the effectiveness of \Policys by varying the number of FC layers, the normalization layer, and the activation function, as shown in \Cref{tab:design}. Since the intermediate outputs ${\bm a}_i \in \mathbb{R}^{K \times 2}$ have two groups, we set the number of groups to two for the group-normalization~\cite{wu2018group} (GN for short). We only change the considered component based on \Method. The accuracy results with an \underline{underline} are higher than the accuracy of \Method. 

\begin{table}[h]
  \centering
  \caption{The accuracy (\%) with various \Policys on Tiny-ImageNet in the default practical setting.}
  \resizebox{\linewidth}{!}{
    \begin{tabular}{l|c|ccc|cc|cc}
    \toprule
     & \Method & \multicolumn{3}{c|}{Number of FC layers} & \multicolumn{2}{c|}{Normalization layer} & \multicolumn{2}{c}{Activation function}\\
    \midrule
    & / & 2 FC & 3 FC & 4 FC & BN & GN & tanh & sigmoid \\
    \midrule
    4-layer CNN & 43.49 & 43.22 & 43.29 & 43.31 & \underline{\textbf{44.13}} & 43.10 & \underline{43.89} & \underline{43.92} \\
    ResNet-18 & 44.18 & \underline{44.50} & \underline{44.36} & 44.11 & 43.70 & 43.25 & 43.49 & \underline{\textbf{44.69}} \\
    \bottomrule
    \end{tabular}}
    \label{tab:design}
\end{table}

The results in \Cref{tab:design} show that we can further improve \Method by using other architectures for the \Policy. Adding more \textbf{FC layers} to process its input improves the test accuracy for ResNet-18 but causes a slight decrease for the 4-layer CNN. The additional parameters introduced for \Method with 1 FC, 2 FC, 3 FC, and 4 FC are 0.527M (million), 0.790M, 1.052M, and 1.315M, respectively. However, the additional computing cost in each iteration introduced by additional FC layers is not worth the little accuracy increase. As for the \textbf{normalization layer}, replacing the LN with the batch-normalization~\cite{ioffe2015batch} (BN) improves 0.64\% test accuracy for the 4-layer CNN. However, it decreases around 0.48\% accuracy for ResNet-18, which also contains BN layers. Similar to LN that normalizes entire ${\bm a}_i$, GN respectively normalizes ${\bm a}_{i, 1}$ and ${\bm a}_{i, 2}$. However, the test accuracy for both the 4-layer CNN and ResNet-18 decreases with the GN layer. As for the activation function, using tanh only increases the accuracy for the 4-layer CNN, while using sigmoid improves the performance for both backbones compared to using ReLU, as the output belongs to $(0, 1)$ is more suitable for outputting a policy. 

\section{Hyperparameter Settings}
\label{sec:hyper}

We use the grid search to find the optimal $\lambda$. Specifically, we perform the grid search in the following search space:
\begin{itemize}
    \item $\lambda$: {$0$, $0.1$, $1$, $5$, $10$}
\end{itemize}

In this paper, we set $\lambda=5$ for the 4-layer CNN and $\lambda=1$ for the ResNet-18 and the fastText, respectively.

\section{Data Distribution Visualization}

Here, we show visualizations of the data distributions (including training and test data) in the image and text tasks.

\begin{figure}[H]
	\centering
	\subfigure[$\beta$ = 0.01]{\includegraphics[width=0.36\linewidth]{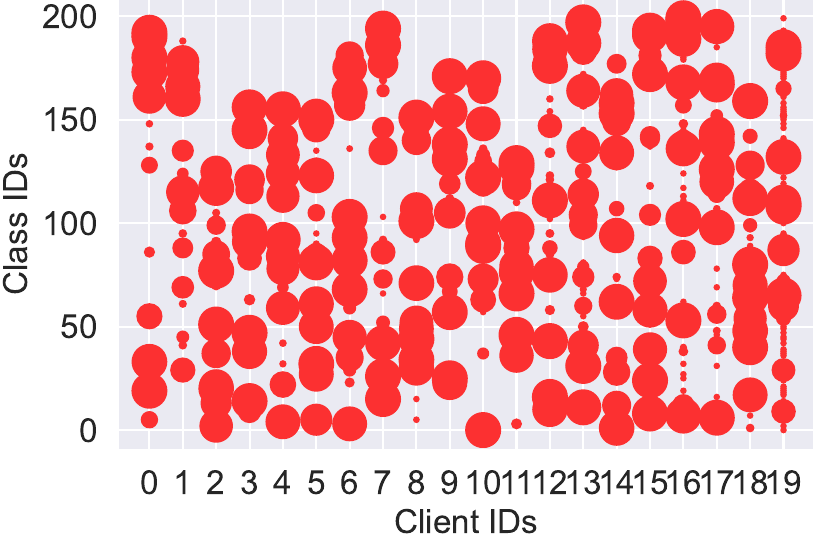}}
    \hfill
	\subfigure[$\beta$ = 0.1]{\includegraphics[width=0.31\linewidth]{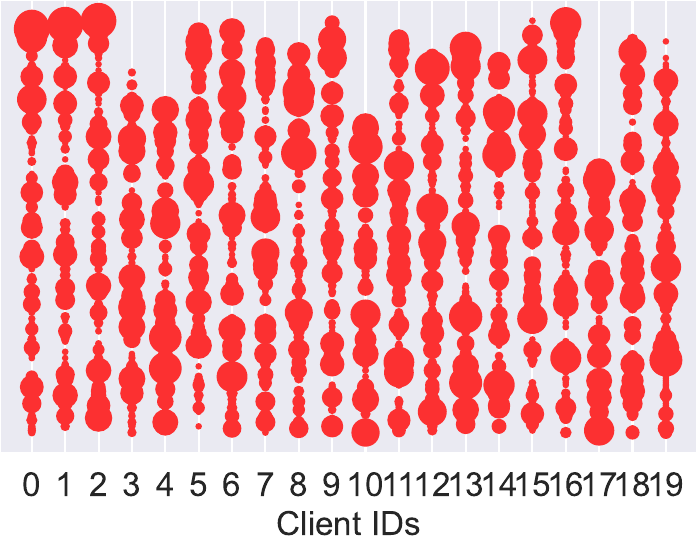}}
	\hfill
	\subfigure[$\beta$ = 0.5]{\includegraphics[width=0.31\linewidth]{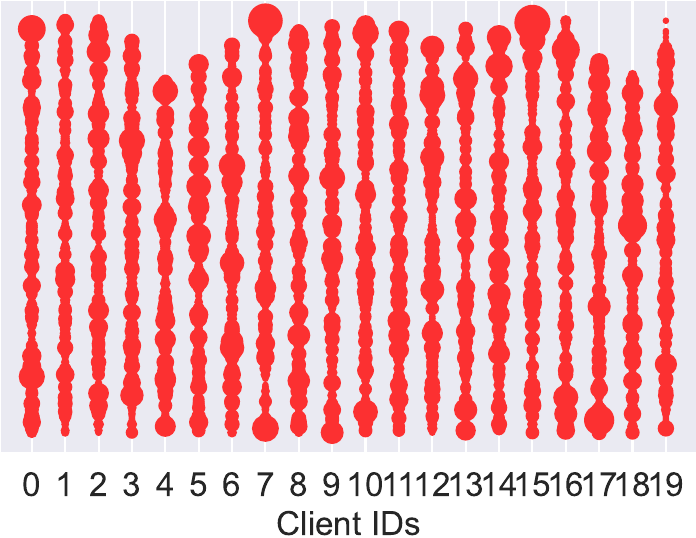}}
	\caption{The data distribution of all clients on Tiny-ImageNet in practical settings with varying $\beta$. The size of a circle means the number of samples.}
	\label{fig:dis_tiny}
\end{figure}

\begin{figure}[H]
	\centering
	\subfigure[10 clients]{\includegraphics[width=0.34\linewidth]{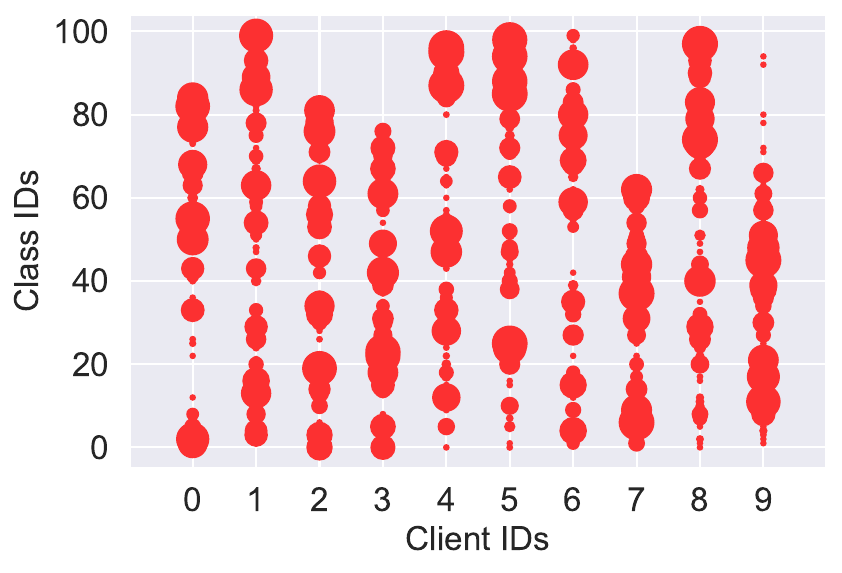}}
	\subfigure[30 clients]{\includegraphics[width=0.64\linewidth]{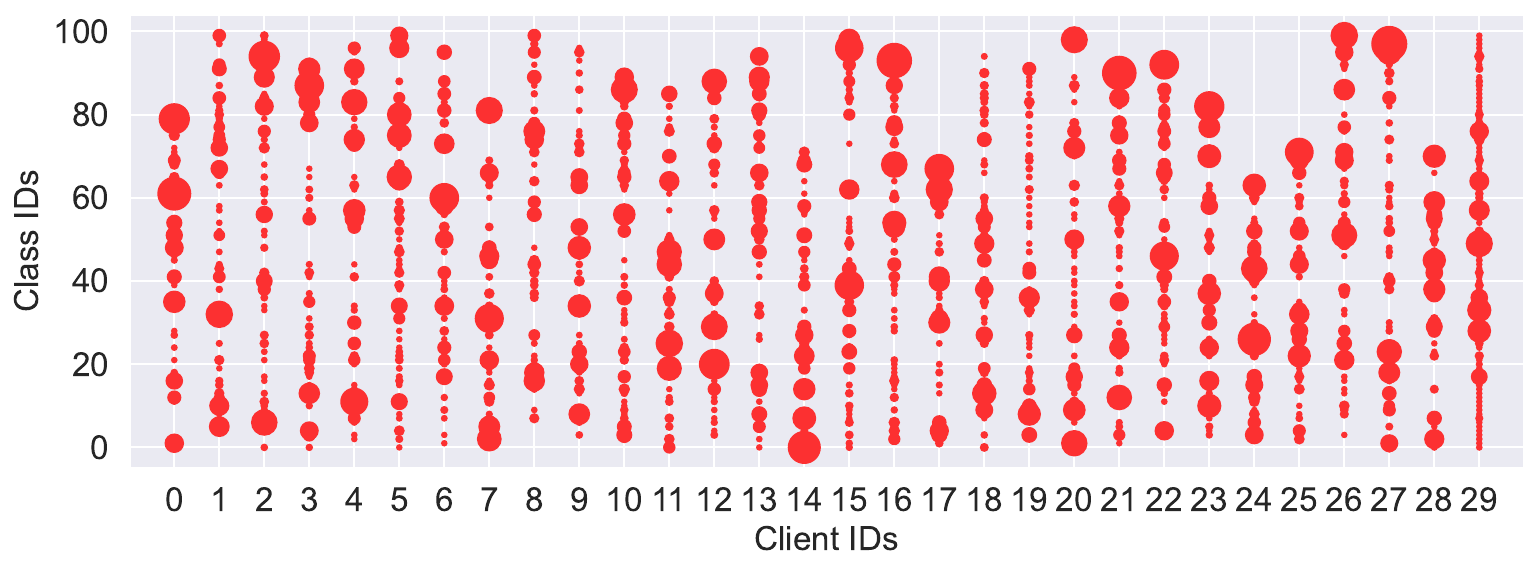}}
	\subfigure[50 clients]{\includegraphics[width=\linewidth]{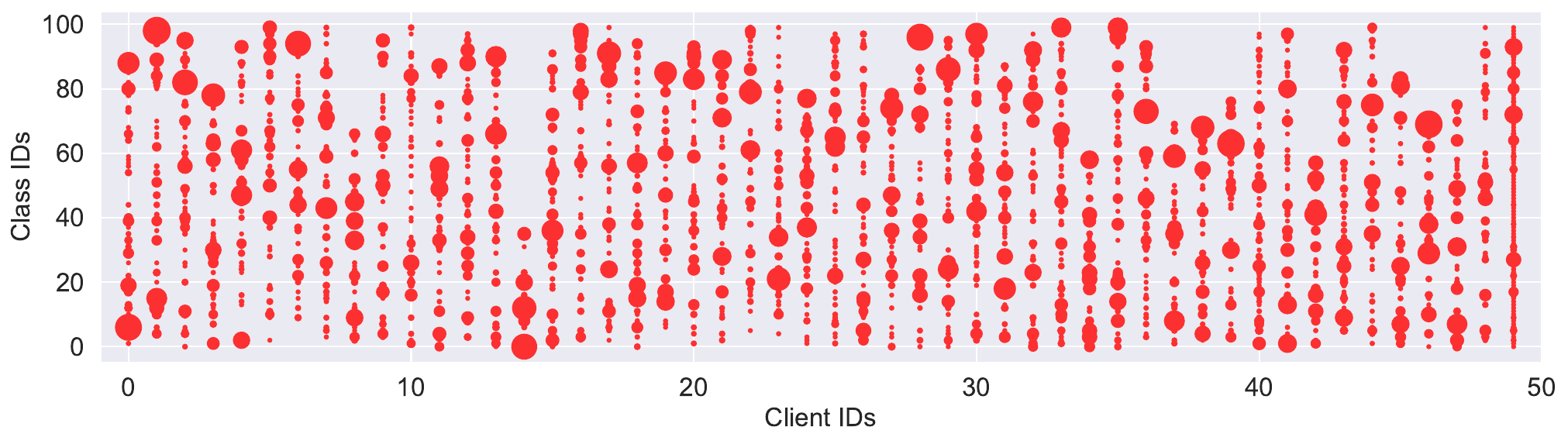}}
	\subfigure[100 clients]{\includegraphics[width=\linewidth]{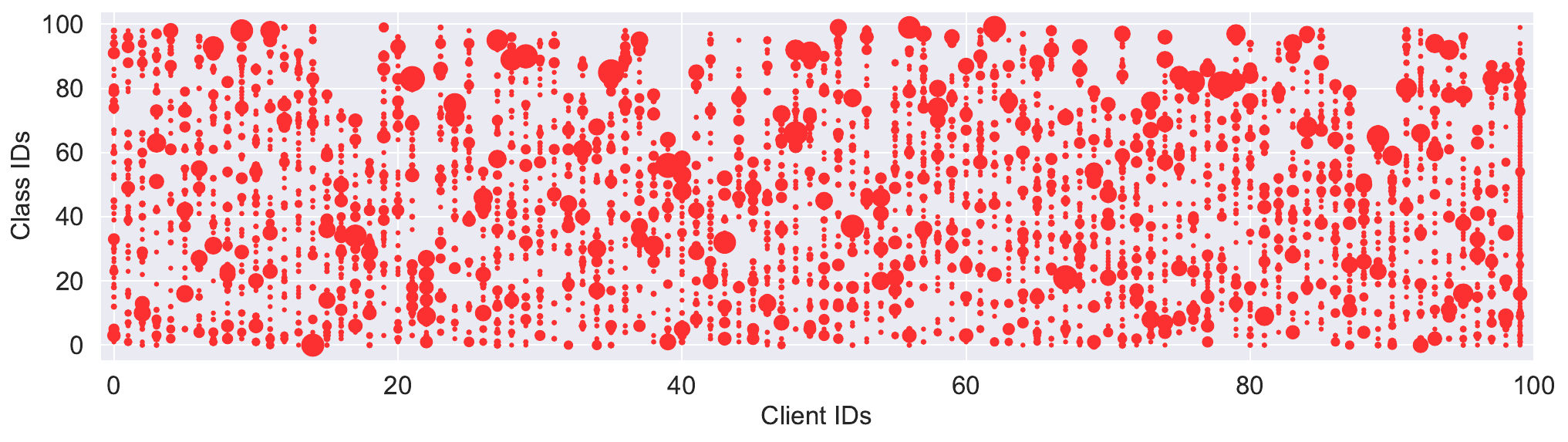}}
	\caption{The data distribution of all clients on Cifar100 in practical settings with 10, 30, 50, and 100 clients, respectively. }
	\label{fig:distribution-50100}
\end{figure}

\begin{figure}[H]
	\centering
	\subfigure[MNIST (pa)]{\includegraphics[width=0.32\linewidth]{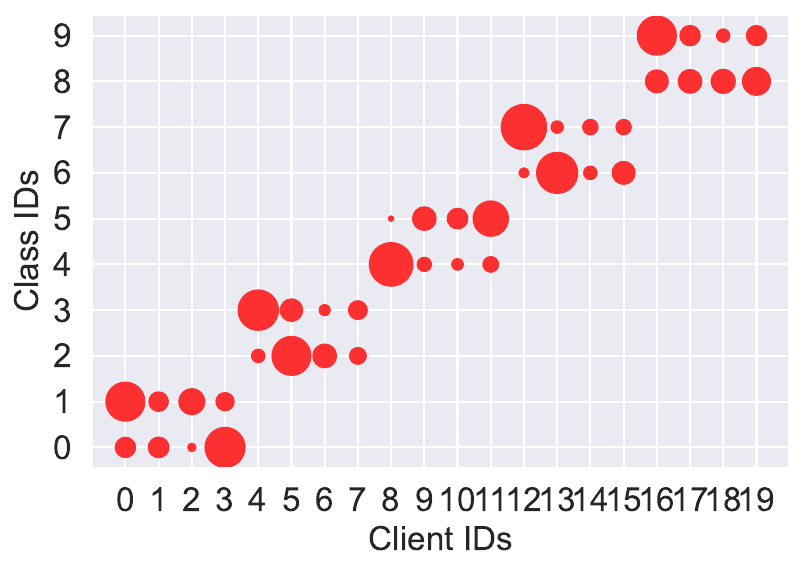}}
    \hfill
	\subfigure[Cifar10 (pa)]{\includegraphics[width=0.32\linewidth]{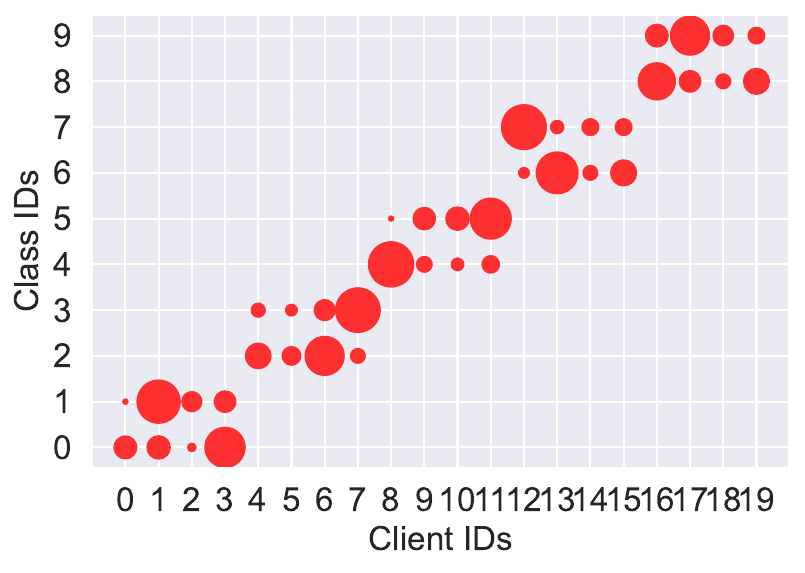}}
	\hfill
	\subfigure[Cifar100 (pa)]{\includegraphics[width=0.32\linewidth]{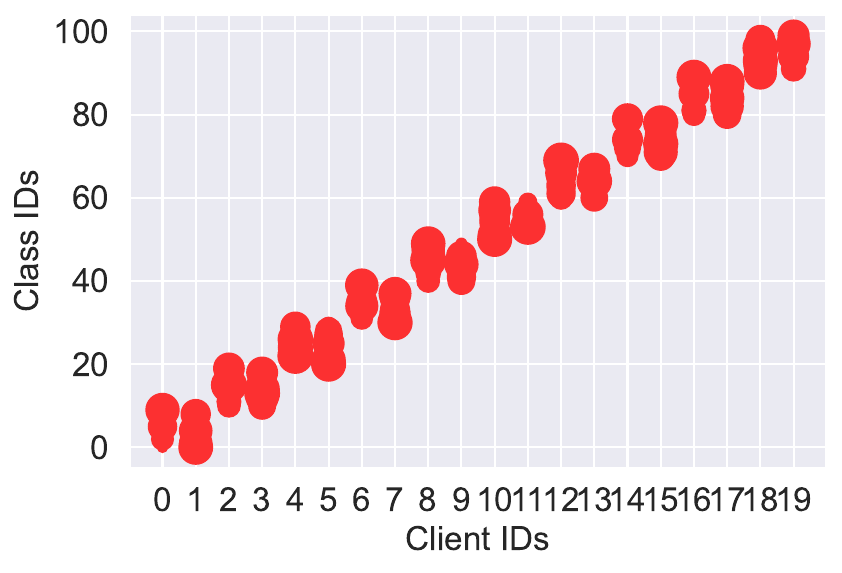}}
	\subfigure[MNIST (pr)]{\includegraphics[width=0.32\linewidth]{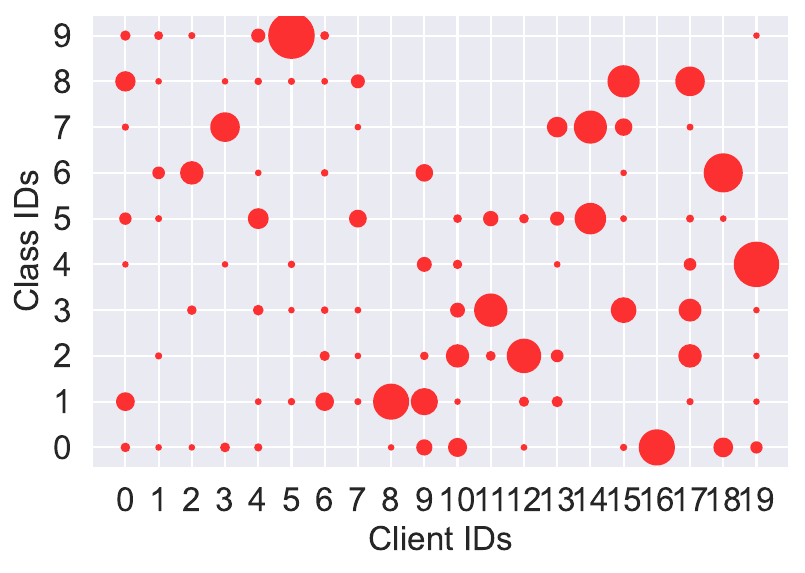}}
    \hfill
	\subfigure[Cifar10 (pr)]{\includegraphics[width=0.32\linewidth]{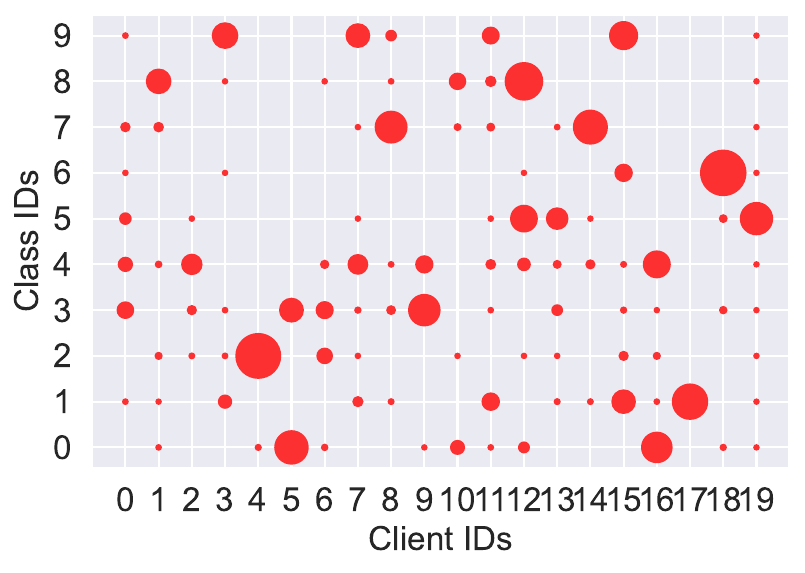}}
	\hfill
	\subfigure[Cifar100 (pr)]{\includegraphics[width=0.32\linewidth]{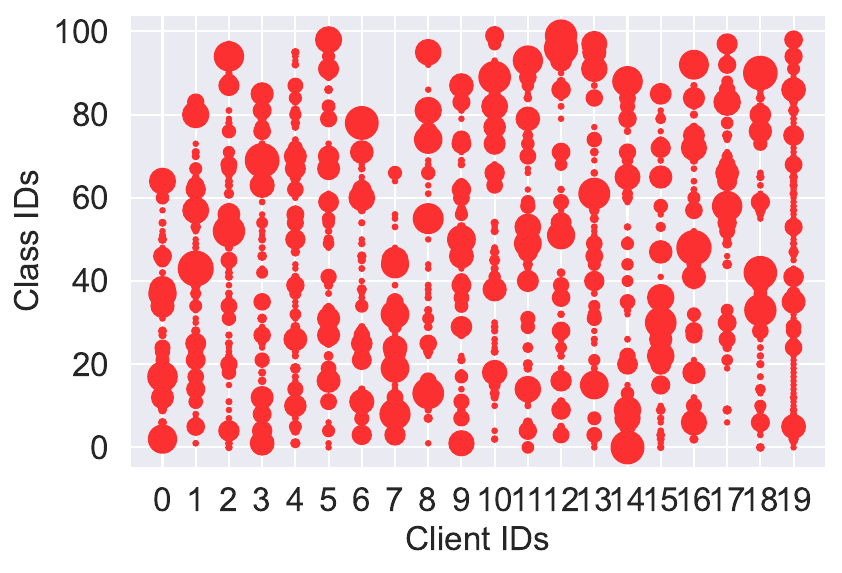}}
	\caption{The data distribution of all clients in the pathological (pa) setting and default practical (pr) setting. }
	\label{fig:distribution-practical}
\end{figure}

\begin{figure}[H]
	\centering
	\subfigure[200 clients]{\includegraphics[width=\linewidth]{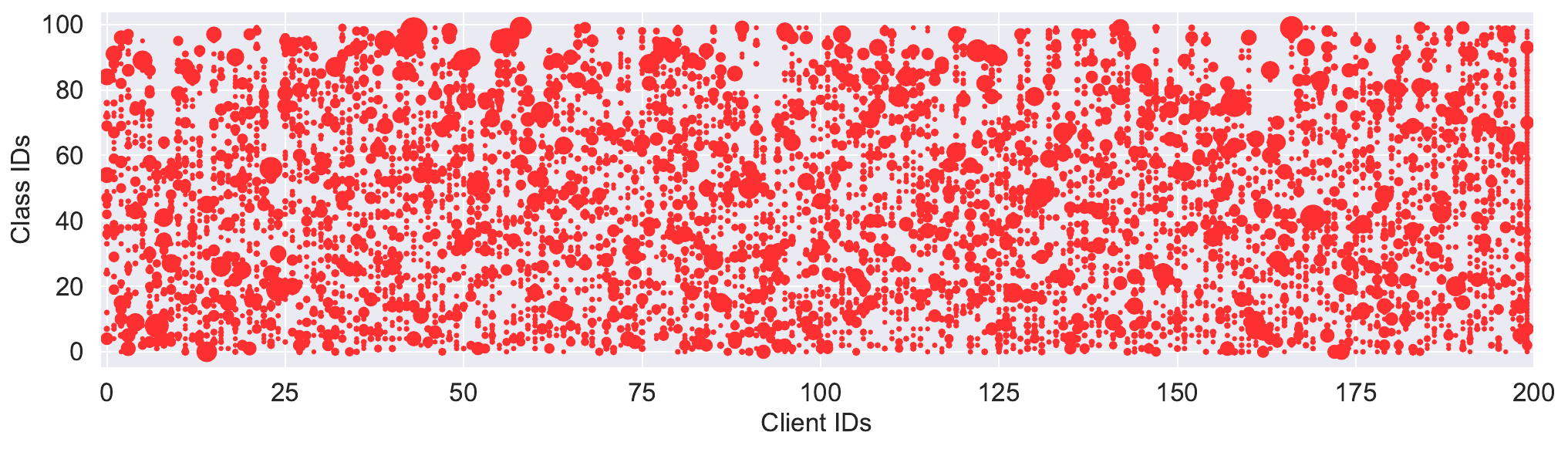}}
	\subfigure[500 clients]{\includegraphics[width=\linewidth]{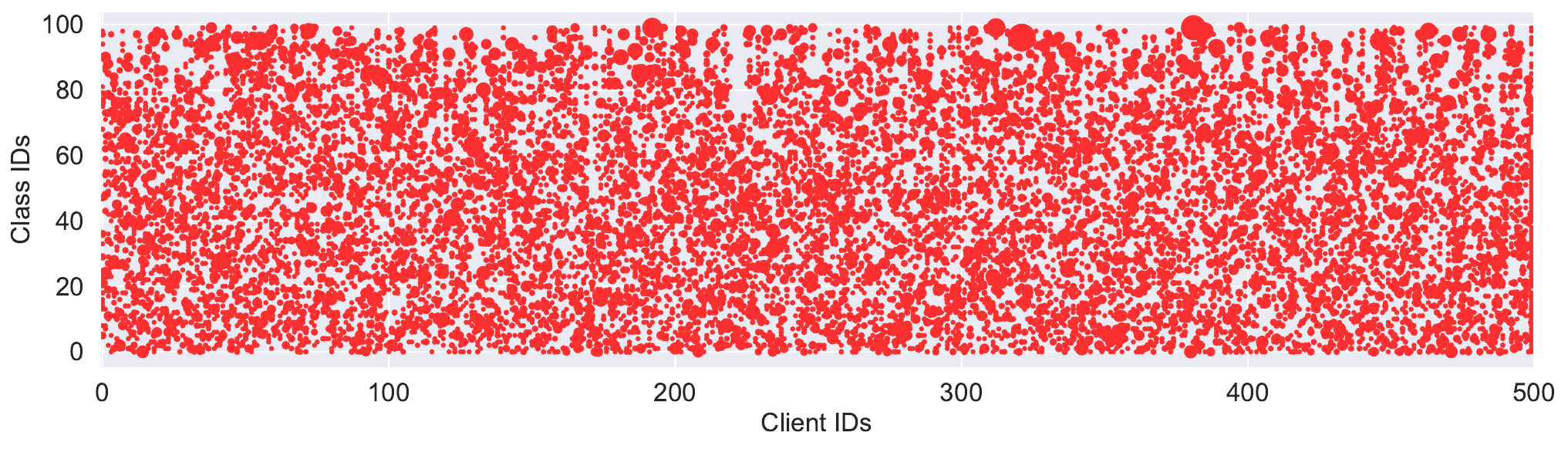}}
	\caption{The data distribution of all clients on Cifar100 in default practical setting with 200 and 500 clients, respectively. }
	\label{fig:distribution-200500}
\end{figure}

\begin{figure}[H]
	\centering
	\subfigure[$\beta$ = 0.1]{\includegraphics[width=0.49\linewidth]{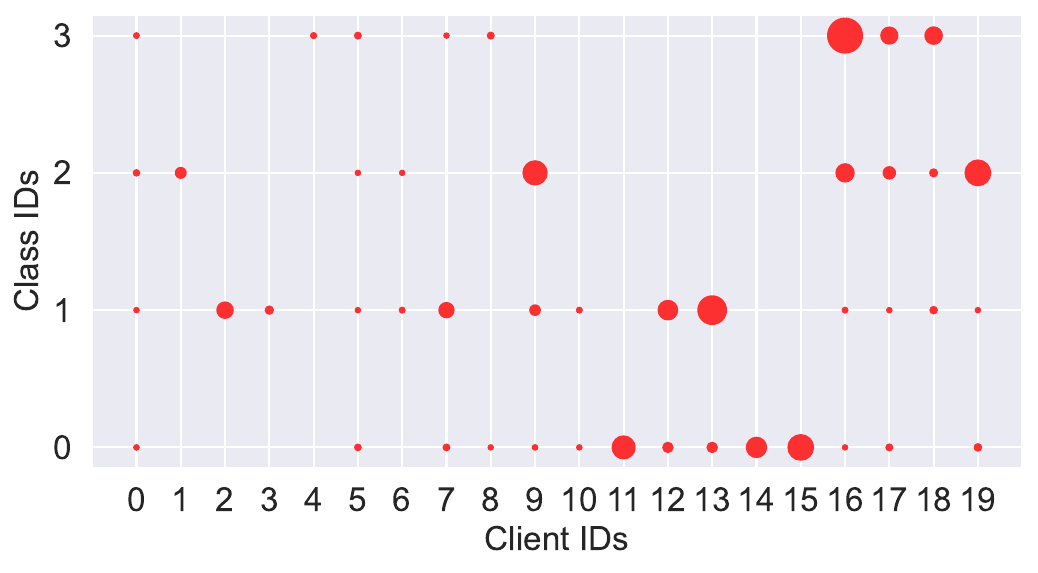}}
	\subfigure[$\beta$ = 1]{\includegraphics[width=0.49\linewidth]{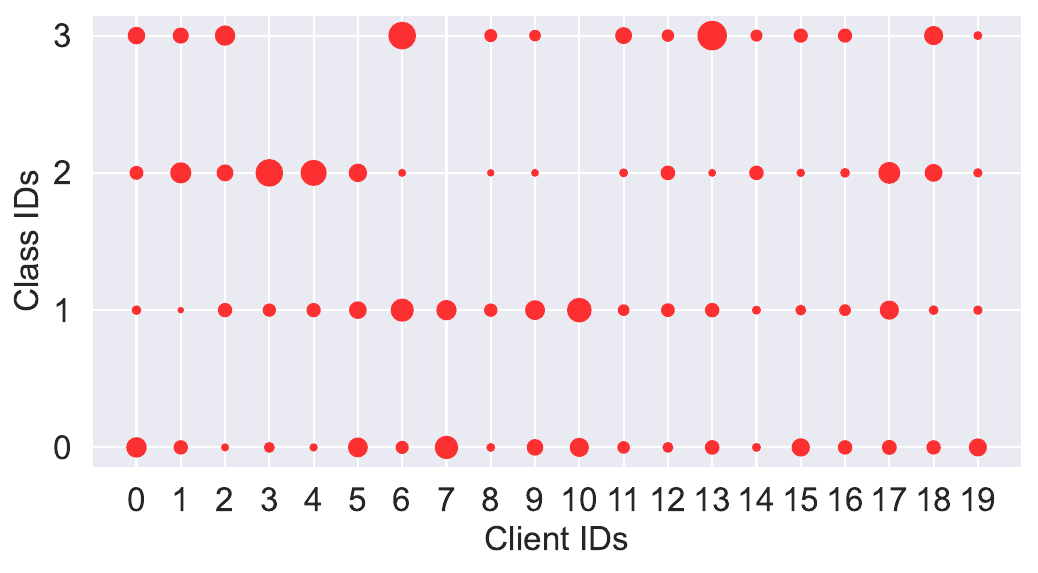}}
	\caption{The data distribution of all clients on AG News in two heterogeneous settings. }
	\label{fig:distribution-ag}
\end{figure}

\end{document}